\documentclass[10pt,twocolumn,letterpaper]{article}

\usepackage{cvpr}
\usepackage{times}
\usepackage{epsfig}
\usepackage{graphicx}
\usepackage{amsmath}
\usepackage{amssymb}

\usepackage[font=small,labelfont=bf]{caption} 
\usepackage{booktabs,siunitx}
\usepackage{tabularx}
\usepackage[usenames, dvipsnames]{xcolor}
\usepackage{multirow,bigdelim}
\usepackage{subcaption}
\usepackage{float}
\usepackage{enumitem}

\graphicspath{{figures/}}

\newcommand{\B}[1]{{\textbf{#1}}}
\newcommand{\SC}[1]{{\textsc{#1}}}
\newcommand{\commentout}[1]{}

\newcommand{\refeqn}[1]{Equation~\ref{#1}}
\newcommand{\reffig}[1]{Figure~\ref{#1}}
\newcommand{\reftbl}[1]{Table~\ref{#1}}
\newcommand{\refsec}[1]{Section~\ref{#1}}

\usepackage{breakcites}
\definecolor{citecolor}{RGB}{34, 139, 34}
\usepackage[pagebackref=true,breaklinks=true,letterpaper=true,colorlinks,citecolor=citecolor,bookmarks=false,hypertexnames=false]{hyperref}

\cvprfinalcopy %

\ifcvprfinal\fi
\begin{document}

\title{Grounded Human-Object Interaction Hotspots from Video}

\author{Tushar Nagarajan\thanks{Work done during internship at Facebook AI Research.}\\
UT Austin\\
{\tt\small tushar@cs.utexas.edu}
\and
Christoph Feichtenhofer\\
Facebook AI Research\\
{\tt\small feichtenhofer@fb.com}
\and
Kristen Grauman\\
Facebook AI Research\\
{\tt\small grauman@fb.com\thanks{On leave from UT Austin (\texttt{grauman@cs.utexas.edu}).}}
}

\maketitle

\begin{abstract}
Learning how to interact with objects is an important step towards embodied visual intelligence, but existing techniques suffer from heavy supervision or sensing requirements.  
We propose an approach to learn human-object interaction ``hotspots" directly from video.
Rather than treat affordances as a manually supervised semantic segmentation task, our approach learns about interactions by watching videos of real human behavior and anticipating afforded actions.
Given a novel image or video, our model infers a spatial hotspot map indicating how an object would be manipulated in a potential interaction---even if the object is currently at rest.
Through results with both first and third person video, we show the value of grounding affordances in real human-object interactions.  Not only are our weakly supervised hotspots competitive with strongly supervised affordance methods, but they can also anticipate object interaction for novel object categories.  
Project page: \url{http://vision.cs.utexas.edu/projects/interaction-hotspots/}
\end{abstract}

\section{Introduction}

Today's visual recognition systems know how objects \emph{look}, but not how they \emph{work}.
Understanding how objects function is fundamental to moving beyond passive perceptual systems (\eg, those trained for image recognition)  to active, embodied agents that are capable of both perceiving and interacting with their environment---whether to clear debris in a search and rescue operation, cook a meal in the kitchen, or even engage in a social event with people.
Gibson's theory of affordances~\cite{gibson1979ecological} provides a way to reason about object function.  It suggests that objects have ``action possibilities" (\eg, a chair affords sitting, a broom affords cleaning), and has been studied extensively in computer vision and robotics in the context of action, scene, and object understanding~\cite{hassanin2018visual}.

However, the abstract notion of ``what actions are possible?'' is only half the story. For example, for an agent tasked with sweeping the floor with a broom, knowing that the broom handle \emph{affords holding} and the broom \emph{affords sweeping} is not enough.  
The agent also needs to know \emph{how} to interact with different objects, including
the best way to grasp the object,
the specific points on the object that need to be manipulated for a successful interaction, %
how the object is used to achieve a goal, and even what it suggests about how to interact with \emph{other} objects. %

\begin{figure}[t!]
\centering
\includegraphics[width=\columnwidth]{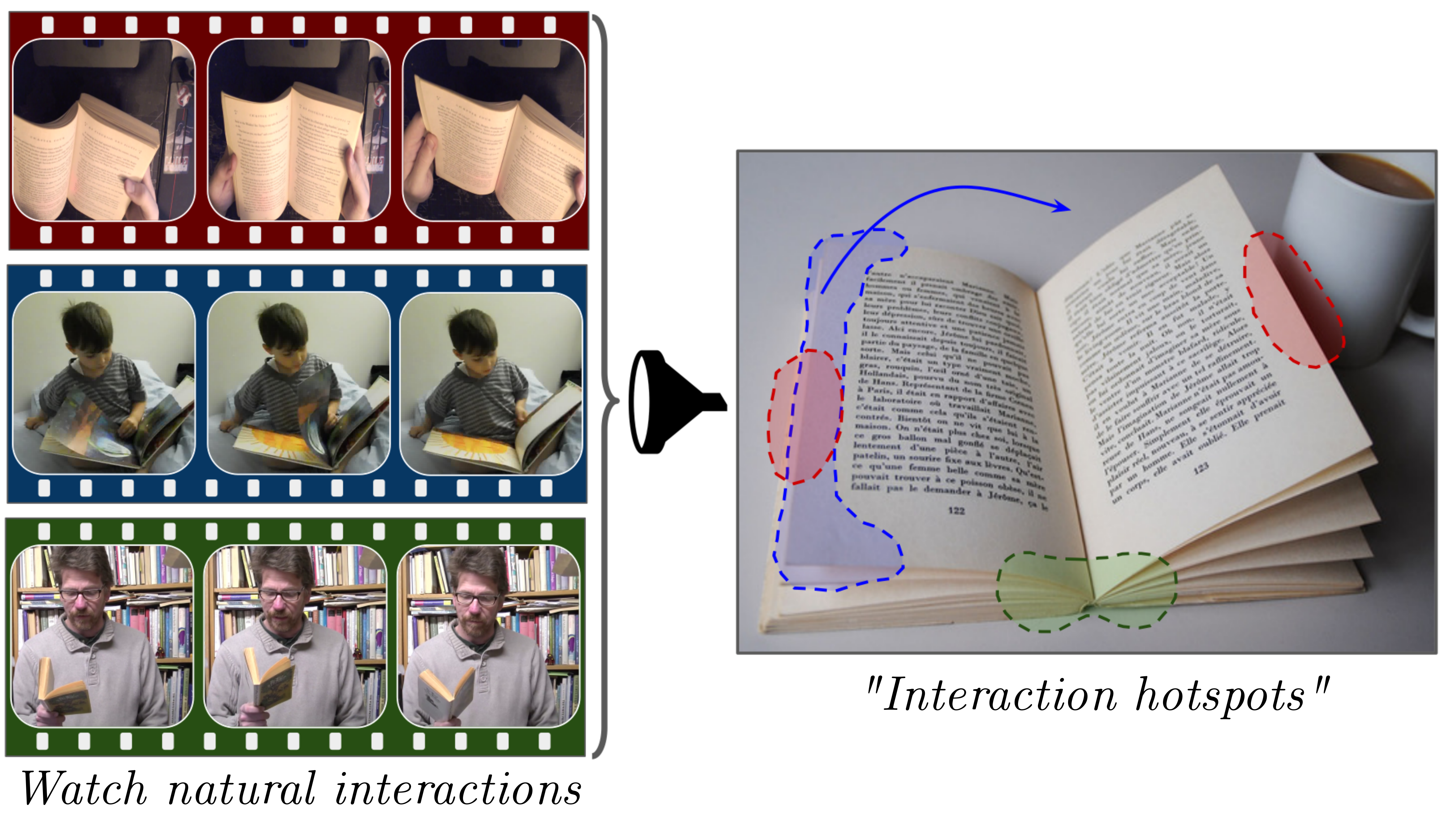}
\caption{%
We propose to learn object affordances directly from videos of people naturally interacting with objects.  The resulting  representation of ``interaction hotspots" is grounded in real human behavior from video, rather than manual image annotations.}
\label{fig:concept}
\end{figure}

Learning how to interact with objects is challenging.  Traditional methods face two key limitations.  First, methods that consider affordances as properties of an object's shape or appearance~\cite{myers2015affordance,grabner2011makes,hermans2011affordance} fall short of modeling actual object use and manipulation.  In particular, learning to segment specified object parts~\cite{nguyen2016detecting,sawatzky2017weakly,myers2015affordance,nguyen2017object} can capture annotators' expectations of what is important, but is detached from real interactions, which are dynamic, multi-modal, and may only partially overlap with part regions (see \reffig{fig:concept}).
Secondly, existing methods are limited by their heavy supervision and/or sensor requirements.  They  assume access to training images with manually drawn masks or keypoints~\cite{roy2016multi,do2017affordancenet,fang2018demo2vec} and some leverage additional sensors like depth~\cite{koppula2014physically,zhu2016inferring,zhu2015understanding} or force gloves~\cite{castellini2011using}, all of which restrict scalability.  Such bottlenecks also deter generalization: exemplars are often captured in artificial lab tabletop environments~\cite{myers2015affordance,koppula2014physically,sawatzky2017weakly} and labeling cost naturally restricts the scope to a narrow set of objects.

In light of these issues, we propose to learn affordances that are \emph{grounded} in real human behavior directly from videos of people naturally interacting with objects, and without any keypoint or mask supervision.
Specifically, we introduce an approach to infer an object's \emph{interaction hotspots}---the spatial regions most relevant to human-object interactions. Interaction hotspots link \emph{inactive} objects at rest not only to the actions they afford, but also to \emph{how} they afford them.
By learning hotspots directly from video, we sidestep issues stemming from manual annotations, avoid imposing part labels detached from real interactions, and discover exactly how people interact with objects in the wild.

Our approach works as follows. First, we use videos of people performing everyday activities to learn an action recognition model %
that can recognize the array of afforded actions when they are \emph{actively in progress} in novel videos.
Then, we introduce an anticipation model to distill the information from the video model, such that it can estimate how a static image of an \emph{inactive} object transforms during an interaction.  In this way, we learn to anticipate the plausible interactions for an object at rest (\eg, perceiving ``cuttable" on the carrot, despite no hand or knife being in view).
Finally, we propose an activation mapping technique tailored for fine-grained object interactions to derive interaction hotspots from the anticipation model.
Thus, given a new image, we can hypothesize interaction hotspots for an object, even if it is not being actively manipulated.

We validate our model on two diverse video datasets: OPRA~\cite{fang2018demo2vec} and EPIC-Kitchens~\cite{damen2018scaling}, spanning hundreds of object and action categories, with videos from both first and third person viewpoints. Our results show that with just weak action and object labels for training video clips, our interaction hotspots can predict object affordances more accurately than prior weakly supervised approaches, with relative improvements up to 25\%. 
Furthermore, we show that our hotspot maps can anticipate object function for novel object classes that are never seen during training, and that our model's learned representation encodes functional similarities between objects that go beyond appearance features.

\vspace{0.1in}
\noindent In summary, we make the following contributions:
\begin{itemize}[leftmargin=*]
    \itemsep0em 
    \item We present a framework that integrates action recognition, a novel anticipation module, and feature localization to learn object affordances directly from video, without manually annotated segmentations/keypoints.
    \item We propose a class activation mapping strategy tailored for fine-grained object interactions that can learn high resolution, localized activation maps.
    \item Our approach predicts affordances more accurately than prior weakly supervised methods---and even competitively with strongly supervised methods---and can anticipate object interaction for novel object classes unobserved in the training video. %
\end{itemize}

\section{Related Work}

\vspace{0.05in}
\noindent\textbf{Visual Affordances}.
The theory of affordances \cite{gibson1979ecological}, originally from work in psychology, has been adopted to study several tasks in computer vision~\cite{hassanin2018visual}. 
In action understanding, affordances provide context for action anticipation~\cite{koppula2016anticipating,rhinehart2016learning,zhou2016cascaded} and  help learn stronger action recognition models~\cite{koppula2013learning}.
In scene understanding, they help decide \textit{where} in a scene a particular action can be performed \cite{savva2014scenegrok,grabner2011makes,wang2017binge,delaitre2012scene}, learn scene geometry~\cite{gupta20113d,fouhey2014people}, or understand social situations~\cite{chuang2017learning}.
In object understanding, affordances help model object function and interaction~\cite{stark2008functional,yao2010grouplet,zhu2015understanding}, and have been studied jointly with hand pose/configuration~\cite{kjellstrom2011visual,thermos2017deep,castellini2011using} and object motion~\cite{gupta2007objects,gupta2009observing}. %

The choice of affordance representation varies significantly in these tasks, spanning across human pose, trajectories of objects, sensorimotor grasps, and 3D scene reconstructions. Often, this results in specialized hardware and heavy sensor requirements (\eg, force gloves, depth cameras). We propose to %
automatically learn appropriate representations for visual affordances  directly from RGB video of human-object interactions.

\vspace{0.05in}
\noindent\textbf{Grounded Affordances}.
Pixel-level segmentation of object parts~\cite{sawatzky2017weakly,myers2015affordance,nguyen2017object} is a common affordance representation, for which 
supervised semantic segmentation frameworks are the typical %
approach~\cite{myers2015affordance,roy2016multi,nguyen2017object,do2017affordancenet}. These segmentations convey high-level information about object function, but rely on manual mask annotations to train---which are not only costly, but can also give an unrealistic view of how objects are actually used.  Unlike our approach, such methods are ``ungrounded" in the sense that the annotator declares regions of interest on the objects outside of any interaction context.

Representations that are grounded in human behavior have also been explored. In images, human body pose serves as a proxy for object affordance to reveal 
modes of interaction with musical instruments~\cite{yao2010grouplet,yao2013discovering} or likely object interaction regions~\cite{chaoyeh-ijcv2016}.  
Given a video, methods can parse 3D models 
to estimate physical concepts (velocity, force, etc.) in order to categorize object interactions~\cite{zhu2016inferring,zhu2015understanding}.  For instructional video, methods explore ways to extract object states~\cite{alayrac2017joint}, modes of object interaction~\cite{damen2014you}, interaction regions~\cite{fang2018demo2vec}, or the anticipated trajectory of an object given a person's skeleton pose~\cite{koppula2014physically}. %

We introduce a new approach for learning affordance ``heatmaps" grounded in human-object interaction, as
derived directly from watching real-world videos of people using the objects.
Our model differs from other approaches in two main ways. First, no prior about interaction in the form of human pose, hand position, or 3D object reconstruction is used. All information about the interactions is learned directly from video. Second, %
rather than learn from manually annotated ground truth masks or keypoints~\cite{myers2015affordance,roy2016multi,nguyen2017object,do2017affordancenet,sawatzky2017weakly,sawatzky2017adaptive,fang2018demo2vec}, 
our model uses only coarse action labels for video clips to guide learning.

\vspace{0.05in}
\noindent\textbf{Video anticipation}. 
Predicting future frames in videos has been studied extensively in computer vision~\cite{ranzato2014video,mathieu2015deep,vondrick2017generating,liang2017dual,srivastava2015unsupervised,villegas2017learning,walker2017pose,oh2015action,xue2016visual,vondrick2016generating,jayaraman2018time}. Future prediction has been applied to action anticipation~\cite{huang2014action,vondrick2015anticipating,koppula2016anticipating,rhinehart2018forecast}, active-object forecasting~\cite{furnari2017next}, %
and to guide demonstration learning in robotics~\cite{finn2016unsupervised,finn2017deep,ebert2017self}.
In contrast to these works, we 
devise a novel anticipation task---learning object interaction affordances from video. 
Rather than predict future frames or action labels, our model anticipates 
correspondences between \emph{inactive} objects (at rest, and not interacted with) and \emph{active} objects (undergoing interaction) in feature space, which we then use to estimate affordances.

\section{Approach}

Our goal is to learn ``interaction hotspots": characteristic %
object regions that anticipate and explain human-object interactions (see \reffig{fig:concept}).  
Conventional approaches for learning affordance segmentation  only address part of this goal. Their manually annotated segmentations are expensive to obtain, do not capture the dynamics of object interaction, and are based on the annotators' notion of importance, which does not always align with real object interactions.
Instead of relying on such segmentations as proxies for interaction, we train our model on a more direct source---videos of people naturally interacting with objects. We contend that such videos contain much of the cues necessary to piece together \emph{how} objects are interacted with.

Our approach consists of three steps.  First, we train a video action classifier to recognize each of the afforded actions (\refsec{sec:action_clf}).
Second, we introduce a novel anticipation model that maps static images of the inactive object to its afforded actions (\refsec{sec:distill}).
Third, we propose an activation mapping technique in the joint model tailored for discovering interaction hotspots on objects, without any keypoint or segmentation supervision (\refsec{sec:hotspot_modifications}).
Given a static image of a novel object, we use the learned model to extract its \emph{hotspot hypotheses} (\refsec{sec:inference}). Critically, the model can infer hotspots even for objects unseen during training, and regardless of whether the object is actively being interacted with in the test image.

\subsection{Learning Afforded Actions from Video} \label{sec:action_clf}

Our key insight is to learn about object interactions from video.  In particular, our approach learns to predict afforded actions across a span of objects, then translates the video cues to static images of an object at rest.
In this way, without explicit region labels and without direct estimation of physical contact points, we learn to anticipate object use.
Throughout, we use the term ``active" to refer to the object when it is involved in an interaction (\ie, the status during training) and ``inactive" to refer to an object at rest with no interaction (\ie, the status during testing).

Let $\mathcal{A}$ denote the set of all afforded actions (\eg, \emph{pourable}, \emph{pushable}, \emph{cuttable}), and let $\mathcal{O}$ denote the set of object categories (\eg, \emph{pan}, \emph{chair}, \emph{blender}), each of which affords one or more actions in $\mathcal{A}$. During training, we have video clips containing various combinations of afforded actions and objects.  %

\begin{figure*}[htb!]
\centering
\includegraphics[width=\linewidth]{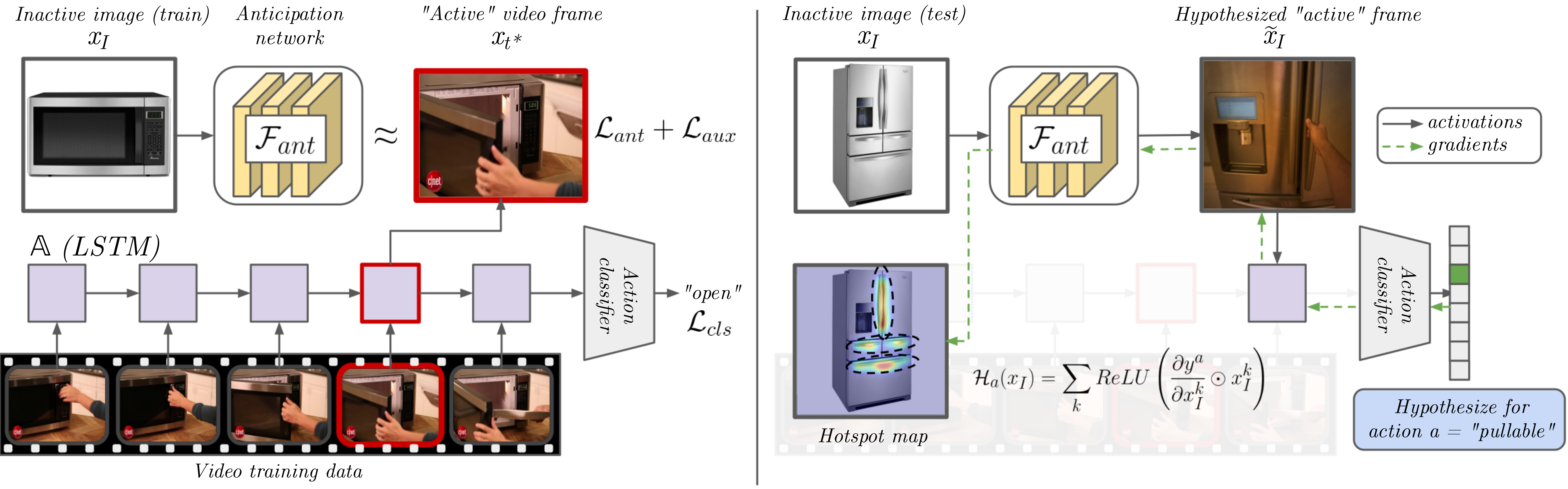}
\caption{\textbf{Illustration of our framework for training (left) and testing (right)}. \textbf{Left panel}: The two components of our model ---the video action classifier (Sec.~\ref{sec:action_clf}) and the anticipation module with its associated losses (Sec.~\ref{sec:distill} and~\ref{sec:hotspot_modifications})---are jointly trained to predict the action class in a video clip while building an affordance-aware internal representation for objects. \textbf{Right panel}: Once trained, our model generates ``interaction hotspot" maps for a novel \emph{inactive} object image (top left fridge image).  It first hallucinates features that would occur for the object \emph{if it were active} (top right photo), then derives gradient-weighted attention maps over the original image, for each action. %
Our method can infer hotspots even for novel object categories unseen in the training video; for example, learning about opening microwaves helps anticipate how to open the fridge. Note that $x_I$, $\widetilde{x}_I$ are in feature space, not pixel space.
}
\label{fig:model}
\end{figure*}

First, we train a video-classification model to predict which afforded action occurs in a video clip. %
For a video of $T$ frames $\mathcal{V}=\{f_1, ..., f_T\}$ and afforded action class $a$, we encode each frame using a convolutional neural network backbone to yield $\{x_1, ..., x_T\}$. Each $x_t$ is a tensor with $d$ channels, each with an $n \times n$ spatial extent, 
with $d$ and $n$ determined by the specific backbone used.\footnote{For example, our experiments use a modified ResNet~\cite{he2016deep} backbone, resulting in $d=2048$ and $n=28$, or a $2048\times 28\times 28$ feature per frame.}  These features are then spatially pooled to obtain a $d$-dimensional vector per frame:
\begin{equation}
    g_t(x_t) = P(x_t) ~~~~~~~~\textrm{for $t=1,\dots,T$}, \\
    \label{eq:action_clf}
\end{equation}
where $P$ denotes the L2-pooling operator.  %
We describe the choice to use this over traditional average pooling in \refsec{sec:hotspot_modifications}.
We further aggregate the frame-level features over time,
\begin{equation}
    h_*(\mathcal{V}) = \mathbb{A}(g_1,\dots,g_T),
\end{equation}
where $\mathbb{A}$ is a \emph{video aggregation module} that combines the frame features of a video into an aggregate feature $h_*$ for the whole video. 
In our experiments, we use a long short-term memory (LSTM) recurrent neural network~\cite{hochreiter1997long} for $\mathbb{A}$. We note that our framework is general and other video classification architectures (\eg, 3D ConvNets) can be used.

The aggregate video feature $h_*$ is then fed to a linear classifier to predict the afforded action, which is trained using cross-entropy loss $\mathcal{L}_{cls}(h_*, a)$.
Once trained, this model can predict which action classes are observed in a video clip of arbitrary length. 
See \reffig{fig:model} (left) for the architecture. 

Note that the classifier's predictions are \emph{object category-agnostic}, since we train it to recognize an afforded action across instances of any object category that affords that action.  In other words, the classifier knows $|\mathcal{A}|$ total actions, not $|\mathcal{A}| \times |\mathcal{O}|$; it recognizes $\emph{pourable}+X$ as one entity, as opposed to $\emph{pourable}+\emph{cup}$ and $\emph{pourable}+\emph{bowl}$ separately. This point is especially relevant once we leverage the model below to generalize hotspots to unfamiliar object classes.

\subsection{Anticipation for Inactive Object Affordances}\label{sec:distill}

So far, we have a video recognition pipeline that can identify occurrences of the afforded actions on sequences with active human-object interactions.
This model alone would focus on ``active" cues directly related to the action being performed (\eg, hands approaching an object), but would not respond strongly to \emph{inactive} instances---static images of objects that are at rest and not being interacted with. 
In fact, prior work demonstrates that these two incarnations of objects are visually quite different, to the point of requiring distinct object detectors, \eg, to recognize both open and closed microwaves~\cite{deva-adl-2012}.

We instead aim for our system to learn about object affordances by watching %
video of people handling objects, then mapping that knowledge to %
novel inactive object photos/frames. 
To bridge this gap, we introduce a distillation-based anticipation module $\mathcal{F}_{ant}$ that transforms the embedding of an inactive object $x_I$, where no interaction is occurring, into its active state where it is being interacted with:
\begin{equation}
    \widetilde{x}_I = \mathcal{F}_{ant}(x_I).
\end{equation} 
See \reffig{fig:model}, top-left.
In experiments we consider two avenues for obtaining the inactive object training images $x_I$: %
inactive frames from a training sequence showing the object before an interaction starts, or catalog photos of the object shown at rest.
During training, the anticipation module is guided by the video action classifier, which selects the appropriate \emph{active state} from a given video as the frame $x_{t^*}$ at which the LSTM is maximally confident of the true action: 
\begin{equation}
   t^* = \operatorname*{arg\,min}_{t \in 1..T} \mathcal{L}_{cls}(\mathbb{A}(g_1, ..., g_t), a),
\end{equation}
where $a$ is the true afforded action label, and $\mathcal{L}_{cls}$ is again the cross-entropy loss for classification using the aggregated hidden state at each time $t$.

We then define a feature matching loss between (a) the anticipated active state for the inactive object and (b) the active state selected by the classifier network for the training sequence. This loss requires the anticipation model to hypothesize a grounded representation of what an object would look like during interaction, according to the actual training video:
\begin{equation} \label{eqn:anticipation}
   \mathcal{L}_{ant}(x_I, x_{t^*}) = ||P(\widetilde{x}_I) - P(x_{t^*}) ||_2.
\end{equation}

Additionally, we make sure that the newly anticipated representation $\widetilde{x}_I$ is predictive of the afforded action and compatible with our video classifier, by using it to classify the afforded action after a single step of the LSTM, resulting in an auxiliary classification loss $\mathcal{L}_{aux}(h_1(\widetilde{x}_I),a)$.

Overall, these components allow our model to estimate what a static inactive object may potentially look like---in feature space---if it were to be interacted with. They provide a crucial %
link between classic action recognition and affordance learning.  As we will define next,  activation mapping through $\mathcal{F}_{ant}$ then provides information about what spatial locations on the original static image are most strongly correlated to how it would be interacted with.

\subsection{Interaction Hotspot Activation Mapping}\label{sec:hotspot_modifications}

At test time, given an inactive object image $q$, 
our goal is to infer the \emph{interaction hotspot} maps $\mathcal{H}_a$, for all $a \in \mathcal{A}$, each of which is an $H \times W$ matrix summarizing the regions of interest that characterize an object interaction, where $H, W$ denote the height and width of the source image.\footnote{To process a novel video, we simply compute hotspots for each frame.} 
Intuitively, a hotspot map should pick up on the regions of the object that would be manipulated or otherwise transform during the action $a$, indicative of its affordances. Note that there is one map per action $a \in \mathcal{A}$.  

We devise an activation mapping approach to go from inactive image embeddings $x_I$ to interaction predictions $\mathcal{F}_{ant}(x_I)$, and finally to hotspots, tailoring it for discovering our hotspot maps. 
For a particular inactive image embedding $x_I$ and an action $a$, we compute the gradient of the score for the action class with respect to each channel of the embedding. These gradients are used to weight individual spatial activations in each channel, acting as an attention mask over them. The positive components of the resulting tensor are retained and accumulated over all channels in the input embedding to give the final hotspot map $\mathcal{H}_a(x_{I})$ for the action class: 
\begin{equation}
    \mathcal{H}_a(x_{I}) = \sum_k ReLU\left(\frac{\partial y^a}{\partial  x_I^k} \odot x_I^k\right),
\label{eq:gradcam1}
\end{equation}
where $x_I^k$ is the $k^{th}$ channel of the input frame embedding and $\odot$ is the element-wise multiplication operator. 
This is meaningful only when the gradients are not spatially uniform (\eg, not if $x_I$ is average pooled for classification). We use L2-pooling to ensure that spatial locations produce gradients as a function of their activation magnitudes.

Next, we address the spatial resolution. The reduced spatial resolution from repeatedly downsampling features in the typical ResNet backbone is reasonable for classification, but is a bottleneck for learning interaction hotspots. We set the spatial stride of the last two residual stages to 1 (instead of 2), and use a dilation for its filters. This increases the spatial resolution by 4$\times$ to $n=28$, allowing our heatmaps to capture finer details. 

\begin{figure}[t!]
\centering
\includegraphics[width=\linewidth]{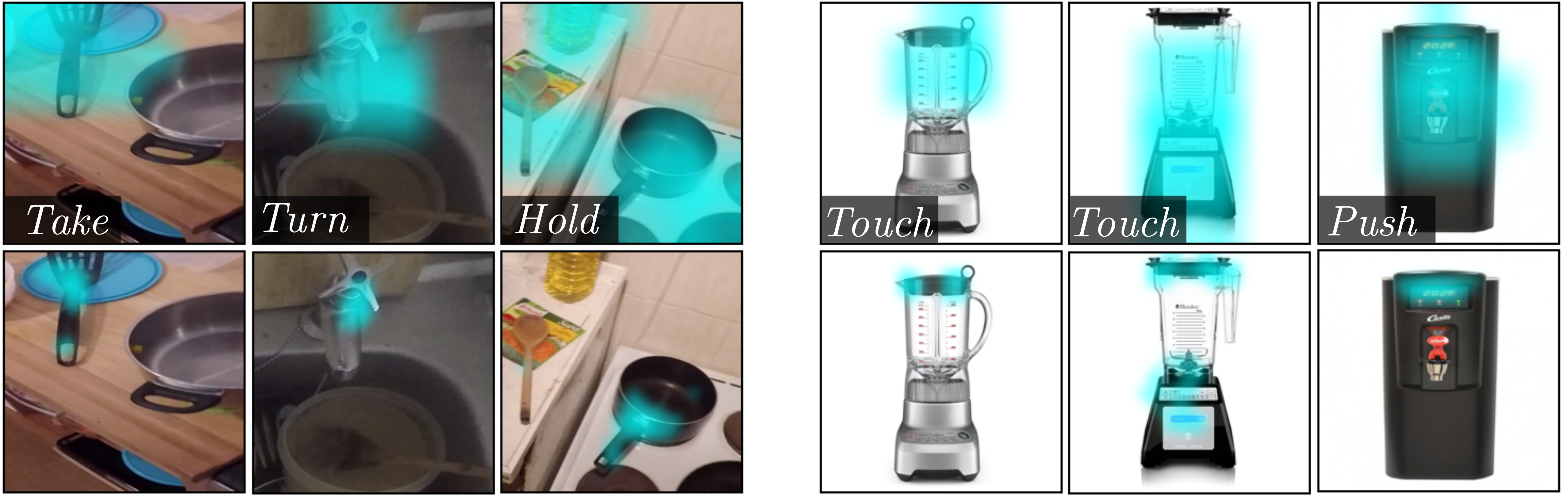}\vspace*{-0.1in}
\caption{\textbf{Our method (bottom) vs.~traditional action recognition+Grad-CAM (top).} Our model generates localized affordance-relevant activations, which a standard action recognition model is unable to do. }
\label{fig:before-after}
\end{figure}

Our technique is related to other feature visualization methods~\cite{springenberg2015striving,zhou2016learning,selvaraju2017grad}. %
However, we use a reduced stride and L2 pooling to make sure that the gradients themselves are spatially localized, and like~\cite{springenberg2015striving}, we do not spatially average gradients---we directly weight activations by them and sum over channels. This is in contrast to GradCAM~\cite{zhou2016learning,selvaraju2017grad} which produces maps that are useful for coarse object localization, but insufficient for interaction hotspots due to their low spatial resolution and diffuse global responses.  Compared to simply applying GradCAM to an action recognition LSTM (\reffig{fig:before-after}, top row), our model produces interaction hotspots that are significantly richer (bottom row). These differences are precisely due to both our anticipation distillation model trained jointly with the recognition model (Sec.~\ref{sec:distill}), as well as the activation mapping strategy above. We provide a quantitative comparison in Sec.~\ref{sec:exp}.

\subsection{Training and Inference} \label{sec:inference}

During training (\reffig{fig:model}, left), we generate embeddings $\{x_1, ..., x_T\}$ for each frame of a video $\mathcal{V}$. These are passed through $\mathbb{A}$ to generate the video embedding $h_*$, and then through a classifier to predict the afforded action label $a$. At the same time the inactive object image embedding $x_I$ is computed and used to train the anticipation model to predict its \emph{active} state $\widetilde{x}_I$. 
The complete loss function for each training instance is:
\begin{equation} \label{eqn:total_loss}
    \mathcal{L}(\mathcal{V},\mathcal{I}, a) = 
    \lambda_{cls} \mathcal{L}_{cls}+
    \lambda_{ant} \mathcal{L}_{ant} +
    \lambda_{aux} \mathcal{L}_{aux},
\end{equation}
where the $\lambda$ terms control the weight of each component of the loss, and $\mathcal{I}$ denotes the inactive object image.

For inference on an inactive test image (\reffig{fig:model}, right), we first generate its image embedding $x_I$. Then, we hypothesize its \emph{active} interaction embedding $\widetilde{x}_I$, and use it to predict the afforded action scores.
Finally, using \refeqn{eq:gradcam1} we generate $|\mathcal{A}|$ heatmaps over $x_I$, one for each afforded action class. This stack of heatmaps are the \emph{interaction hotspots}.
Note that we produce activation maps for the original inactive image $x_I$, not for the hypothesized active output $\widetilde{x}_I$, \ie, we propagate gradients \emph{through} the anticipation network as well. Not doing so produces activation maps that are inconsistent with the input image, which hurts performance (see ablation study in Supp).

We stress that interaction hotspots are predictable even for unfamiliar objects.  By training the afforded actions across object category boundaries, the system learns the general properties of appearance \emph{and} interaction that characterize affordances.  %
Hence, our approach can anticipate, for example, how an unfamiliar kitchen device might be used because it has learned how a variety of other objects operate. Similarly, heatmaps can be hallucinated for novel action-object pairs that have not been seen in training (\eg, ``cut" using a spatula in \reffig{fig:opra-qual}, bottom row).  %

Please see Supp. for complete implementation details.
\section{Experiments}  \label{sec:exp}

Our experiments on interaction hotspots explore their ability to describe affordances of objects, to generalize to anticipate affordances of unfamiliar objects, and to encode functional similarities between object classes.

\begin{table}
\centering
\resizebox{\columnwidth}{!}{
\begin{tabular}{llll}\toprule
                                                            & Supervision Source                          & Type     & $N$ \\ \midrule
\SC{egogaze}~\cite{huang2018predicting}                     & Recorded eye fixations                      & Weak     & 60k \\
\SC{saliency}~\cite{pan2017salgan,mlnet2016,deepgaze2016}   & Manual saliency labels                      & Weak     & 10k \\
\SC{ours}                                                   & Action, object labels                       & Weak     & 20k \\ \midrule
\SC{img2heatmap}                                            & Manual affordance keypoints                 & Strong   & 20k \\

\multirow{2}{*}{\SC{demo2vec}~\cite{fang2018demo2vec}}
                                                            & Manual affordance keypoints,                & \multirow{2}{*}{Strong}    & \multirow{2}{*}{20k} \\ 
                                                            & action labels                               &          &     \\ \bottomrule
\end{tabular}
}
\caption{\textbf{Supervision source and type for all methods.} Our method learns interaction hotspots \emph{without} strong supervision like annotated segmentation/keypoints. $N$ is the number of instances.}%
\label{tbl:supervision_sources}
\end{table}

\vspace{0.05in}
\noindent\textbf{Datasets}. \label{sec:datasets}
We use two datasets:
\begin{itemize}[leftmargin=*]
\item \textbf{OPRA}~\cite{fang2018demo2vec} contains videos of product reviews of appliances (\eg, refrigerators, coffee machines) collected from YouTube. Each instance is a short video demonstration $\mathcal{V}$ of a product's feature (\eg, pressing a button on a coffee machine) paired with a static image $\mathcal{I}$ of the product, an interaction label $a$ (\eg, ``pressing"), and a manually created affordance heatmap $\mathcal{M}$ (\eg, highlighting the button on the static image). There are $\sim$16k training instances of the form $(\mathcal{V}, \mathcal{I}, a, \mathcal{M})$, spanning 7 actions. %

\item \textbf{EPIC-Kitchens}~\cite{damen2018scaling} contains unscripted, egocentric videos of activities in a kitchen. Each clip $\mathcal{V}$ is annotated with action and object labels $a$ and $o$ (\eg, cut tomato, open refrigerator) along with a set of bounding boxes $\mathcal{B}$ (one per frame) for objects being interacted with. There are $\sim$40k training instances of the form $(\mathcal{V}, a, o, \mathcal{B})$, spanning 352 objects and 125 actions. We crowd-source annotations for ground-truth heatmaps $\mathcal{M}$ %
resulting in 1.8k annotated instances over 20 action and 31 objects (see Supp.~for details).

\end{itemize}

The two video datasets span diverse settings.  OPRA has third person videos, where the person and the product being reviewed are clearly visible, and covers a small number of actions and products. %
EPIC-Kitchens has first-person videos of unscripted kitchen activities and a much larger vocabulary of actions and objects; the person is only partially visible when they manipulate an object.
Together, they provide good variety and difficulty to evaluate the robustness of our model.\footnote{Other affordance segmentation datasets~\cite{myers2015affordance,nguyen2016detecting} have minimal vocabulary overlap with OPRA/EPIC classes, and hence do not permit evaluation for our setting, since we learn from video.} For both datasets, our model uses only the action labels as supervision, and an \emph{inactive} image for our anticipation loss $\mathcal{L}_{ant}$. We stress that (1) the annotated heatmap $\mathcal{M}$ is used \emph{only} for evaluation, and (2) the ground truth is well-aligned with our objective, since annotators were instructed to watch an interaction video clip to decide what regions to annotate for an object's affordances.  %

While OPRA comes with an image $\mathcal{I}$ of the \emph{exact} product associated with each video instance, EPIC does not. Instead, we crop out inactive objects from frames using the provided bounding boxes $\mathcal{B}$, and randomly select one that matches the object class label in the video. To account for the appearance mismatch, in place of the L2 loss in \refeqn{eqn:anticipation} we use a triplet loss,
which uses ``negatives" to ensure that inactive objects of the correct class can anticipate active features better than incorrect classes (see Supp.~for details).

\subsection{Interaction Hotspots as Grounded Affordances} \label{sec:grounded-affordances}

In this section, we evaluate two things: 1) How well does our model learn object affordances? and 2) How well can it infer possible interactions for unfamiliar objects? For this, we train our model on video clips, and generate hotspot maps on \emph{inactive} images where the object is at rest.

\begin{table*}[t]
\resizebox{2\columnwidth}{!}{
\centering
\begin{tabular}{rl|ccc|ccc|c@{\hskip 0.1in}|ccc|ccc|}
\multicolumn{2}{c}{}                     &          \multicolumn{3}{c}{OPRA}                    &           \multicolumn{3}{c}{EPIC} &\multicolumn{1}{c}{}&        \multicolumn{3}{c}{OPRA}                    &           \multicolumn{3}{c}{EPIC}                   \\
\cmidrule{3-8} \cmidrule{10-15}
&                                        & KLD $\downarrow$ & SIM $\uparrow$ & AUC-J $\uparrow$ & KLD $\downarrow$ & SIM $\uparrow$ & AUC-J $\uparrow$ && KLD $\downarrow$ & SIM $\uparrow$ & AUC-J $\uparrow$ & KLD $\downarrow$ & SIM $\uparrow$ & AUC-J $\uparrow$ \\
\cmidrule{2-8} \cmidrule{10-15}
& \SC{center bias}                       & 11.132           & 0.205          & 0.625            & 10.660           & 0.222          & 0.634            && 6.281            & 0.244          & 0.680            & 5.910            & 0.277          & 0.699            \\
\ldelim[{6}{1mm}[{\rotatebox[origin=c]{90}{WS}}]
& \SC{lstm+grad-cam}                     & 8.573            & 0.209          & 0.620            & 6.470            & 0.257          & 0.626            && 5.405            & 0.259          & 0.644           & 4.508           & 0.255          & 0.664             \\
& \SC{egogaze} \cite{huang2018predicting}& 2.428            & 0.245          & 0.646            & 2.241            & 0.273          & 0.614            && 2.083            & 0.278          & 0.694           & 1.974           & 0.298          & 0.673             \\
& \SC{mlnet} \cite{mlnet2016}            & 4.022            & 0.284          & 0.763            & 6.116            & 0.318          & 0.746            && 2.458            & 0.316          & 0.778           & 3.221           & 0.361          & 0.799             \\     
& \SC{deepgazeII} \cite{deepgaze2016}    & 1.897            & 0.296          & 0.720            & 1.352            & 0.394          & 0.751            && 1.757            & 0.318          & 0.742           & 1.297           & 0.400          & 0.793             \\     
& \SC{salgan} \cite{pan2017salgan}       & 2.116            & 0.309          & 0.769            & 1.508            & 0.395          & 0.774            && 1.698            & 0.337          & 0.790           & 1.296           & \B{0.406}      & 0.808             \\     
& \SC{ours}                              & \B{1.427}        & \B{0.362}      & \B{0.806}        & \B{1.258}        & \B{0.404}      & \B{0.785}        && \B{1.381}        & \B{0.374}      & \B{0.826}       & \B{1.249}       & 0.405          & \B{0.817}         \\     
\cmidrule{2-8} \cmidrule{10-15}

\ldelim[{2}{1mm}[{\rotatebox[origin=c]{90}{SS}}]
& \SC{img2heatmap}                       & 1.473            & 0.355          & 0.821            & 1.400            & 0.359          & 0.794            && 1.431            & 0.362          & 0.820           & 1.466           & 0.353          & 0.770             \\ 
& \SC{demo2vec} \cite{fang2018demo2vec}  & 1.197            & 0.482          & 0.847            & --               & --             & --               && --               & --             & --              & --              & --             & --                \\ 
\cmidrule{2-8} \cmidrule{10-15}
\end{tabular}
}
\vskip -0.1in
\subcaptionbox*{\textbf{Grounded affordance prediction}}[.7\linewidth]{}  \hskip -0.7in
\subcaptionbox*{\textbf{Generalization to novel objects}}[.3\linewidth]{} \vskip -0.1in

\caption{\textbf{Interaction hotspot prediction results on OPRA and EPIC}. \textbf{Left:} Our model outperforms other weakly supervised (WS) methods in all metrics, and approaches the performance of strongly supervised (SS) methods \emph{without} the privilege of heatmap annotations during training. \textbf{Right:} Not only does our model generalize to new \emph{instances}, but it also accurately infers interaction hotspots for novel object \emph{categories} unseen during training. The proposed hotspots generalize on an object-function level. Values are averaged across three splits of object classes. ($\uparrow$/$\downarrow$ indicates higher/lower is better.)  \SC{Demo2Vec}~\cite{fang2018demo2vec} is available only on OPRA and only for seen classes.}

\label{tbl:hotspot-eval}
\end{table*}

\vspace{0.05in}
\noindent\textbf{Baselines}. We evaluate our model against several baselines and state-of-the-art models. 
\begin{itemize}[leftmargin=*]
\itemsep0em 
\item \textbf{\SC{Center Bias}} produces a fixed Gaussian heatmap at the center of the image. This is a naive baseline to account for a possible center bias~\cite{mlnet2016,deepgaze2016,pan2017salgan,huang2018predicting}.
\item \textbf{\SC{LSTM+Grad-CAM}} uses an LSTM trained for action recognition with the same weak labels as our method, then applies standard Grad-CAM~\cite{selvaraju2017grad} to get heatmaps.  It has no anticipation model.
\item \textbf{\SC{Saliency}} is a set of baselines that estimate the most salient regions in an image using models trained directly on saliency annotations/eye fixations: \SC{egogaze}~\cite{huang2018predicting}, \SC{mlnet}~\cite{mlnet2016}, \SC{deepgazeII}~\cite{deepgaze2016} and \SC{salgan}~\cite{pan2017salgan}. We use the authors' pretrained models.
\item \textbf{\SC{Demo2Vec}}~\cite{fang2018demo2vec} is a supervised method that
 generates an affordance heatmap using context from a video demonstration of the interaction. We use the authors' pre-computed heatmap predictions.%
\item \textbf{\SC{Img2Heatmap}} is a supervised method that uses a fully convolutional encoder-decoder to predict the affordance heatmap for an image.  It serves as a simplified version of \SC{Demo2Vec} that lacks video context during training.
\end{itemize}
The \SC{Saliency} baselines capture a generic notion of spatial \emph{importance}.  %
They produce a single heatmap for an image, regardless of action class, and as such, are less expressive than our per-action-affordances. They are %
\emph{weakly supervised} in that they are trained for a different task, albeit with strong supervision (heatmaps, gaze points) for that task. 
\SC{Demo2Vec} and \SC{Img2Heatmap} are strongly supervised, and represent more traditional affordance learning techniques that learn affordances from manually labeled images~\cite{myers2015affordance,roy2016multi,nguyen2017object,do2017affordancenet}.
\reftbl{tbl:supervision_sources} summarizes the sources and types of supervision for all methods. Unlike other methods, our model uses only weak class labels during training.

\vspace{0.05in}
\noindent\textbf{Grounded Affordance Prediction}. First we compare the ground truth heatmaps for each interaction to our hotspots and the baselines' heatmaps.  We report error as KL-Divergence, following~\cite{fang2018demo2vec}, as well as other metrics (SIM, AUC-J) from the saliency literature~\cite{bylinskii2018different}.

\reftbl{tbl:hotspot-eval} (Left) summarizes the results.  Our model outperforms all other weakly-supervised methods in all metrics across both datasets. These results highlight that our model can capture sophisticated interaction cues that describe more specialized notions of importance than saliency. 

On OPRA, our model achieves relative improvements of up to 25\% (KLD) compared to the strongest baseline, and it matches one of the strongly supervised baseline methods on two metrics.
On EPIC, our model achieves relative improvements up to 7\% (KLD). EPIC has a much larger, more granular action vocabulary, resulting in fewer and less spatially distinct hotspots. As a result, the baselines that produce redundant heatmaps for all actions artificially benefit on EPIC, though our results remain better. %

\begin{figure}[t!]
\centering
\includegraphics[width=\linewidth]{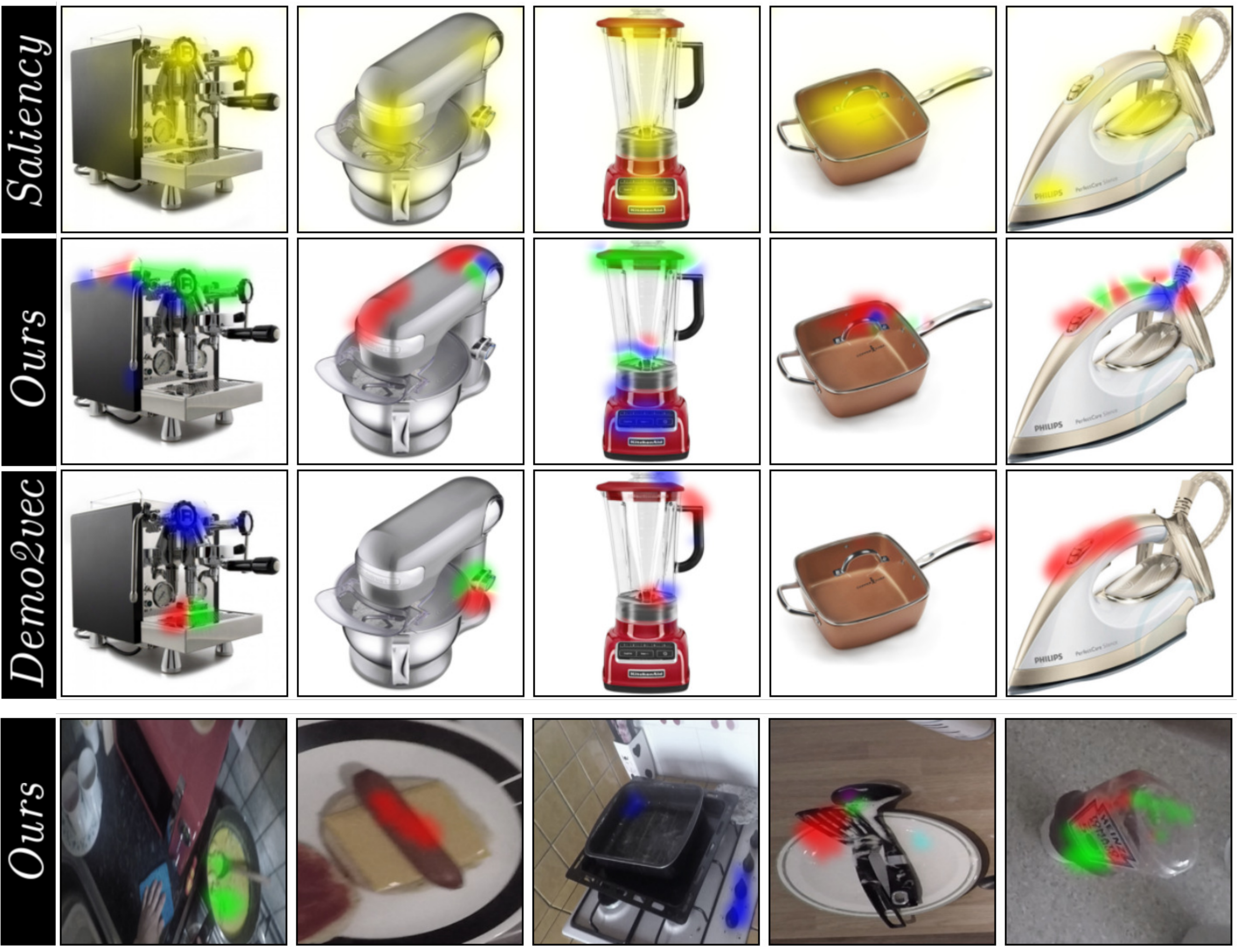}
\vspace*{-0.2in}%
\caption{\textbf{Affordance heatmaps on inactive images.} \textbf{Top:} Predicted affordance heatmaps for \emph{hold}, \emph{rotate}, \emph{push} (red, green, blue) on OPRA. \textbf{Bottom row:} Predicted heatmaps for \emph{cut}, \emph{mix}, \emph{turn-on} (red, green, blue), on EPIC. Our model highlights spatial affordances consistent with how people interact with the objects. Note that \SC{saliency}~\cite{pan2017salgan} produces only a single ``importance" map (yellow).  \textbf{Last column:} failure cases. Best viewed in color.  
} %
\vspace*{-0.1in}
\label{fig:opra-qual}
\end{figure}

The baselines have similar trends across datasets. Consistent with the examples in \reffig{fig:before-after}, the \SC{LSTM+Grad-CAM} baseline in \reftbl{tbl:hotspot-eval} demonstrates  that simply training an action recognition model is clearly insufficient to learn affordances. Our anticipation model allows the system to bridge the (in)active gap between training video and test images, and is crucial for accuracy. 
All saliency methods perform worse than our model, despite that they may accidentally benefit from the fact that kitchen appliances have interaction regions designed to be visually salient (\eg, buttons, handles). 
In contrast to our approach, none of the saliency baselines distinguish between affordances; they produce a single heatmap representing ``important" salient points. To these methods, the blade of a knife is as important to the action ``cutting" as it is to the action ``holding", %
and they are unable to explain objects with multiple affordances.
\SC{img2heatmap} and \SC{demo2vec} generate better affordance heatmaps, but at the cost of strong supervision. Our method actually approaches their accuracy %
without using any manual heatmaps for training.

Please see the Supp.~file for an \textbf{ablation study} that further examines the contributions of each part of our model.  In short, our class activation mapping strategy and propagating gradients all the way through the anticipation model are critical. All elements of the design play a role to achieve our full model's best accuracy.

\reffig{fig:opra-qual} shows example heatmaps for inactive objects. Our model is able to highlight specific object regions that afford actions (\eg, the knobs on the coffee machine as ``rotatable" in column 1) after only watching videos of object interactions. Weakly supervised \SC{Saliency} methods highlight \emph{all} salient object parts in a single map, regardless of the interaction in question. 
In contrast, our model
highlights multiple distinct affordances for an object. To generate comparable heatmaps, \SC{Demo2vec} requires annotated heatmaps for training \emph{and} a set of video demonstrations during inference, whereas our model can hypothesize object functionality without these extra requirements.

\begin{figure}
\centering
\includegraphics[width=1\columnwidth]{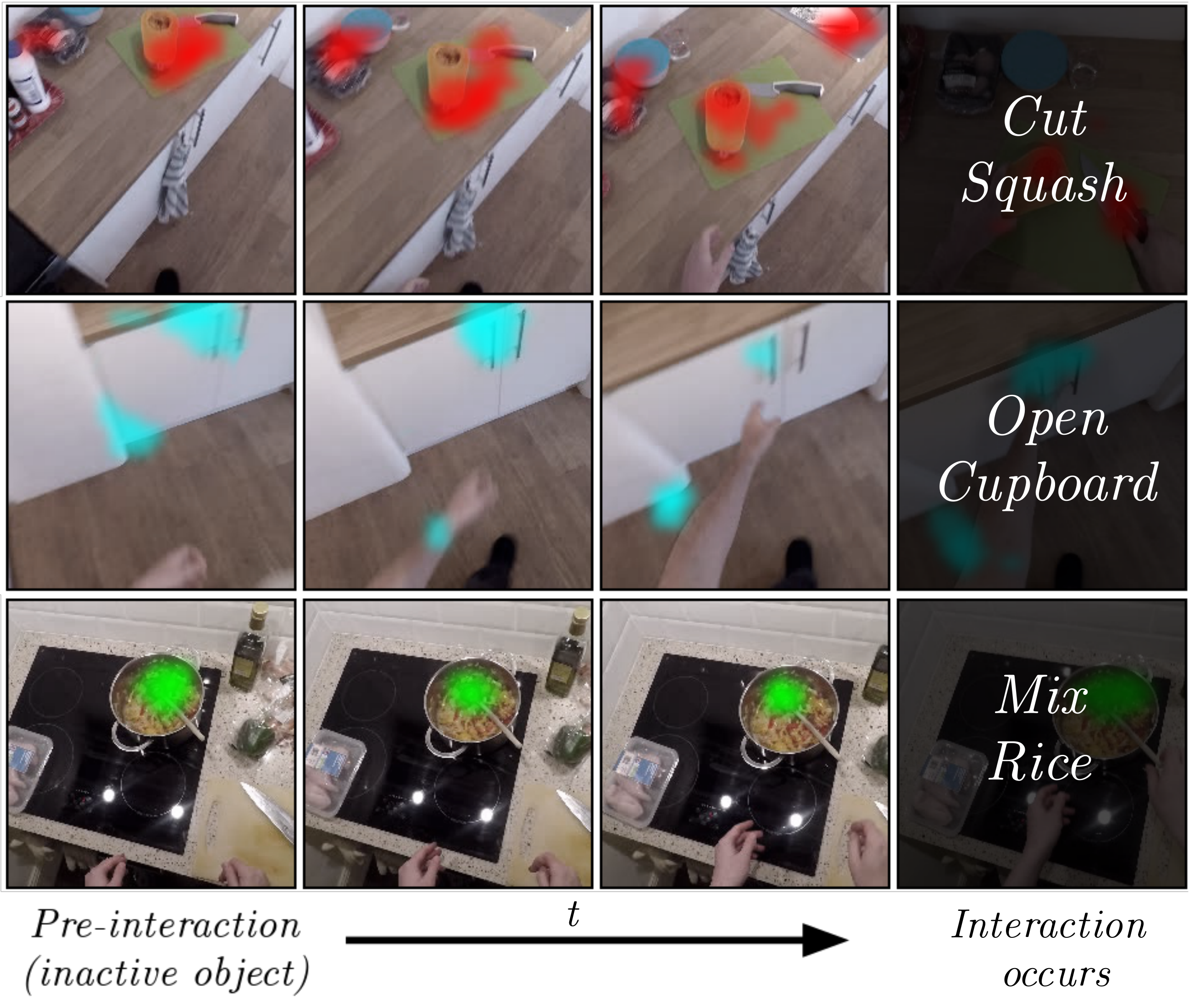}
\vspace*{-0.25in}
\caption{\textbf{Interaction hotspots on EPIC videos of unseen object classes.} %
Our model anticipates spatial interaction regions for inactive objects at rest (first column), \emph{even before} the interaction happens. Critically, the object categories shown in this figure have \emph{not} been seeing during training; our model learns to generalize interaction hotspots.  For example, there are no cupboards or squashes in the training videos, but our method anticipates how these objects would be opened and cut, respectively. } %
\label{fig:epic-video}
\end{figure}

\vspace{0.05in}
\noindent\textbf{Generalization to Novel Objects.} Can interaction hotspots infer how \emph{novel} object categories work?  We next test if our model learns an \emph{object-agnostic} representation for interaction---one that is not tied to object class. This is a useful property for open-world situations where unfamiliar objects may have to be interacted with to achieve a goal. 

We divide the object categories $\mathcal{O}$ into familiar and unfamiliar object categories $\mathcal{O} = \mathcal{O}_f \bigcup \mathcal{O}_u$; familiar ones are those seen with interactions in training video and unfamiliar ones are seen only during testing. We leave out 10/31 objects in EPIC and 9/26 objects in OPRA for our experiments, and divide our video train/test sets along these object splits.  %
We train our model only on clips with the familiar objects from $\mathcal{O}_f$. If our model can successfully infer the heatmaps for novel, unseen objects, it will show that a general sense of object \emph{function} is learned that is not strongly tied to object \emph{identity}. 

\reftbl{tbl:hotspot-eval} (Right) shows the results. We see mostly similar trends as the previous section.  On OPRA, our model outperforms all baselines in all metrics, and is able to infer the hotspot maps for unfamiliar object categories, despite never seeing them during training. 
On EPIC, our method remains the best weakly supervised method.

Qualitative results (\reffig{fig:epic-video}) support our numbers, showing our model applied to video clips from EPIC Kitchens, just before the action occurs. %
Our model---which was never trained on some objects (\eg, cupboard, squash)---is able to anticipate characteristic spatial locations of interactions \emph{before} the interaction occurs. 

\begin{figure}[t]%
\centering
\includegraphics[width=\linewidth]{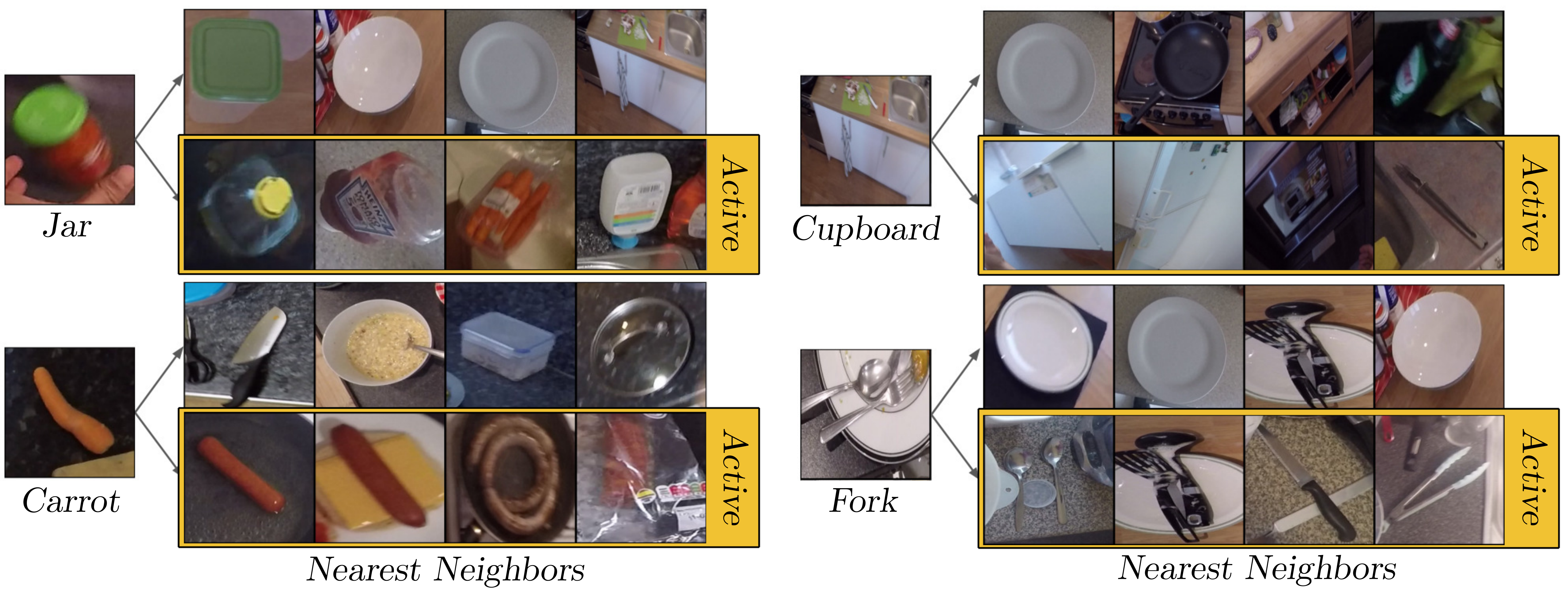}
\vspace*{-0.2in}%
\caption{\textbf{Inactive vs.~active object embeddings}. By hypothesizing potential interactions with objects, our model learns representations that capture functional similarities between objects across object classes, rather than purely appearance-based similarities.}
\label{fig:tsne}
\end{figure}

\subsection{Interaction Hotspots for Functional Similarity} \label{sec:dendo}
Finally, we show how our model encodes functional object similarities in its learned representation for objects. We compare the inactive object embedding space (standard ResNet features) to our predicted active embedding space (output of the anticipation model) by looking at nearest neighbor images in other object classes. %

\reffig{fig:tsne} shows examples.  Neighbors in the inactive object space (top branch) capture typical appearance-based visual similarities that are useful for object categorization---shapes, backgrounds, etc. In contrast, our active object space (bottom branch, yellow box) reorganizes the objects based on \emph{how} they are interacted with.  %
For example, fridges, cupboards, and microwaves, that are swung open in a characteristic way (top right); knives, spatulas, tongs, that are typically held at their handles (bottom right). %
Our model learns representations indicative of functional similarity between objects, despite the objects being visually distinct.
See Supp.~for a clustering visualization on all images.

\section{Conclusion}

We introduced a method to learn ``interaction hotspot" maps---characteristic regions on objects that anticipate and explain object interactions---directly from watching videos of people naturally interacting with objects. 
Our experiments show that these hotspot maps explain object affordances better than other existing weakly supervised models and can generalize to anticipate affordances of unseen objects.  Furthermore, the representation learned by our model goes beyond appearance similarity to encode  functional similarity.  %
In future work, we plan to explore how hotspots might aid %
action anticipation and policy learning for robot-object interaction.

\vspace{0.1in}
{\noindent\textbf{Acknowledgments}: We would like to thank the authors of Demo2Vec~\cite{fang2018demo2vec} for their help with the OPRA dataset. We would also like to thank Marcus Rohrbach and Jitendra Malik for the helpful discussions.}

{\small
\bibliographystyle{ieee}
\bibliography{egbib}

\begin{thebibliography}{10}\itemsep=-1pt

\bibitem{alayrac2017joint}
J.-B. Alayrac, J.~Sivic, I.~Laptev, and S.~Lacoste-Julien.
\newblock Joint discovery of object states and manipulation actions.
\newblock {\em ICCV}, 2017.

\bibitem{bylinskii2018different}
Z.~Bylinskii, T.~Judd, A.~Oliva, A.~Torralba, and F.~Durand.
\newblock What do different evaluation metrics tell us about saliency models?
\newblock {\em TPAMI}, 2018.

\bibitem{castellini2011using}
C.~Castellini, T.~Tommasi, N.~Noceti, F.~Odone, and B.~Caputo.
\newblock Using object affordances to improve object recognition.
\newblock {\em TAMD}, 2011.

\bibitem{chaoyeh-ijcv2016}
C.-Y. Chen and K.~Grauman.
\newblock Subjects and their objects: Localizing interactees for a
  person-centric view of importance.
\newblock {\em IJCV}, 2016.

\bibitem{chuang2017learning}
C.-Y. Chuang, J.~Li, A.~Torralba, and S.~Fidler.
\newblock Learning to act properly: Predicting and explaining affordances from
  images.
\newblock {\em CVPR}, 2018.

\bibitem{mlnet2016}
M.~Cornia, L.~Baraldi, G.~Serra, and R.~Cucchiara.
\newblock {A Deep Multi-Level Network for Saliency Prediction}.
\newblock In {\em ICPR}, 2016.

\bibitem{damen2018scaling}
D.~Damen, H.~Doughty, G.~M. Farinella, S.~Fidler, A.~Furnari, E.~Kazakos,
  D.~Moltisanti, J.~Munro, T.~Perrett, W.~Price, et~al.
\newblock Scaling egocentric vision: The epic-kitchens dataset.
\newblock {\em ECCV}, 2018.

\bibitem{damen2014you}
D.~Damen, T.~Leelasawassuk, O.~Haines, A.~Calway, and W.~W. Mayol-Cuevas.
\newblock You-do, i-learn: Discovering task relevant objects and their modes of
  interaction from multi-user egocentric video.
\newblock In {\em BMVC}, 2014.

\bibitem{delaitre2012scene}
V.~Delaitre, D.~F. Fouhey, I.~Laptev, J.~Sivic, A.~Gupta, and A.~A. Efros.
\newblock Scene semantics from long-term observation of people.
\newblock In {\em ECCV}, 2012.

\bibitem{do2017affordancenet}
T.-T. Do, A.~Nguyen, I.~Reid, D.~G. Caldwell, and N.~G. Tsagarakis.
\newblock Affordancenet: An end-to-end deep learning approach for object
  affordance detection.
\newblock {\em ICRA}, 2017.

\bibitem{ebert2017self}
F.~Ebert, C.~Finn, A.~X. Lee, and S.~Levine.
\newblock Self-supervised visual planning with temporal skip connections.
\newblock {\em CORL}, 2017.

\bibitem{fang2018demo2vec}
K.~Fang, T.-L. Wu, D.~Yang, S.~Savarese, and J.~J. Lim.
\newblock Demo2vec: Reasoning object affordances from online videos.
\newblock In {\em CVPR}, 2018.

\bibitem{finn2016unsupervised}
C.~Finn, I.~Goodfellow, and S.~Levine.
\newblock Unsupervised learning for physical interaction through video
  prediction.
\newblock In {\em NeurIPS}, 2016.

\bibitem{finn2017deep}
C.~Finn and S.~Levine.
\newblock Deep visual foresight for planning robot motion.
\newblock In {\em ICRA}, 2017.

\bibitem{fouhey2014people}
D.~F. Fouhey, V.~Delaitre, A.~Gupta, A.~A. Efros, I.~Laptev, and J.~Sivic.
\newblock People watching: Human actions as a cue for single view geometry.
\newblock {\em IJCV}, 2014.

\bibitem{furnari2017next}
A.~Furnari, S.~Battiato, K.~Grauman, and G.~M. Farinella.
\newblock Next-active-object prediction from egocentric videos.
\newblock {\em JVCIR}, 2017.

\bibitem{gibson1979ecological}
J.~J. Gibson.
\newblock {\em The ecological approach to visual perception: classic edition}.
\newblock Psychology Press, 1979.

\bibitem{grabner2011makes}
H.~Grabner, J.~Gall, and L.~Van~Gool.
\newblock What makes a chair a chair?
\newblock In {\em CVPR}, 2011.

\bibitem{gupta2007objects}
A.~Gupta and L.~S. Davis.
\newblock Objects in action: An approach for combining action understanding and
  object perception.
\newblock In {\em CVPR}, 2007.

\bibitem{gupta2009observing}
A.~Gupta, A.~Kembhavi, and L.~S. Davis.
\newblock Observing human-object interactions: Using spatial and functional
  compatibility for recognition.
\newblock {\em TPAMI}, 2009.

\bibitem{gupta20113d}
A.~Gupta, S.~Satkin, A.~A. Efros, and M.~Hebert.
\newblock From 3d scene geometry to human workspace.
\newblock In {\em CVPR}, 2011.

\bibitem{hassanin2018visual}
M.~Hassanin, S.~Khan, and M.~Tahtali.
\newblock Visual affordance and function understanding: A survey.
\newblock {\em arXiv preprint arXiv:1807.06775}, 2018.

\bibitem{he2016deep}
K.~He, X.~Zhang, S.~Ren, and J.~Sun.
\newblock Deep residual learning for image recognition.
\newblock In {\em CVPR}, 2016.

\bibitem{hermans2011affordance}
T.~Hermans, J.~M. Rehg, and A.~Bobick.
\newblock Affordance prediction via learned object attributes.
\newblock In {\em ICRA: Workshop on Semantic Perception, Mapping, and
  Exploration}, 2011.

\bibitem{hochreiter1997long}
S.~Hochreiter and J.~Schmidhuber.
\newblock Long short-term memory.
\newblock {\em Neural computation}, 1997.

\bibitem{huang2014action}
D.-A. Huang and K.~M. Kitani.
\newblock Action-reaction: Forecasting the dynamics of human interaction.
\newblock In {\em ECCV}, 2014.

\bibitem{huang2018predicting}
Y.~Huang, M.~Cai, Z.~Li, and Y.~Sato.
\newblock Predicting gaze in egocentric video by learning task-dependent
  attention transition.
\newblock {\em ECCV}, 2018.

\bibitem{jayaraman2018time}
D.~Jayaraman, F.~Ebert, A.~A. Efros, and S.~Levine.
\newblock Time-agnostic prediction: Predicting predictable video frames.
\newblock {\em ICLR}, 2019.

\bibitem{kjellstrom2011visual}
H.~Kjellstr{\"o}m, J.~Romero, and D.~Kragi{\'c}.
\newblock Visual object-action recognition: Inferring object affordances from
  human demonstration.
\newblock {\em CVIU}, 2011.

\bibitem{koppula2013learning}
H.~S. Koppula, R.~Gupta, and A.~Saxena.
\newblock Learning human activities and object affordances from rgb-d videos.
\newblock {\em IJR}, 2013.

\bibitem{koppula2014physically}
H.~S. Koppula and A.~Saxena.
\newblock Physically grounded spatio-temporal object affordances.
\newblock In {\em ECCV}, 2014.

\bibitem{koppula2016anticipating}
H.~S. Koppula and A.~Saxena.
\newblock Anticipating human activities using object affordances for reactive
  robotic response.
\newblock {\em TPAMI}, 2016.

\bibitem{krizhevsky2012imagenet}
A.~Krizhevsky, I.~Sutskever, and G.~E. Hinton.
\newblock Imagenet classification with deep convolutional neural networks.
\newblock In {\em NIPS}, 2012.

\bibitem{deepgaze2016}
M.~K{\"u}mmerer, T.~S. Wallis, and M.~Bethge.
\newblock Deepgaze ii: Reading fixations from deep features trained on object
  recognition.
\newblock {\em arXiv preprint arXiv:1610.01563}, 2016.

\bibitem{liang2017dual}
X.~Liang, L.~Lee, W.~Dai, and E.~P. Xing.
\newblock Dual motion gan for future-flow embedded video prediction.
\newblock In {\em ICCV}, 2017.

\bibitem{mathieu2015deep}
M.~Mathieu, C.~Couprie, and Y.~LeCun.
\newblock Deep multi-scale video prediction beyond mean square error.
\newblock {\em ICLR}, 2016.

\bibitem{myers2015affordance}
A.~Myers, C.~L. Teo, C.~Ferm{\"u}ller, and Y.~Aloimonos.
\newblock Affordance detection of tool parts from geometric features.
\newblock In {\em ICRA}, 2015.

\bibitem{nguyen2016detecting}
A.~Nguyen, D.~Kanoulas, D.~G. Caldwell, and N.~G. Tsagarakis.
\newblock Detecting object affordances with convolutional neural networks.
\newblock In {\em IROS}, 2016.

\bibitem{nguyen2017object}
A.~Nguyen, D.~Kanoulas, D.~G. Caldwell, and N.~G. Tsagarakis.
\newblock Object-based affordances detection with convolutional neural networks
  and dense conditional random fields.
\newblock In {\em IROS}, 2017.

\bibitem{oh2015action}
J.~Oh, X.~Guo, H.~Lee, R.~L. Lewis, and S.~Singh.
\newblock Action-conditional video prediction using deep networks in atari
  games.
\newblock In {\em NeurIPS}, 2015.

\bibitem{pan2017salgan}
J.~Pan, C.~C. Ferrer, K.~McGuinness, N.~E. O'Connor, J.~Torres, E.~Sayrol, and
  X.~Giro-i Nieto.
\newblock Salgan: Visual saliency prediction with generative adversarial
  networks.
\newblock {\em arXiv preprint arXiv:1701.01081}, 2017.

\bibitem{deva-adl-2012}
H.~Pirsiavash and D.~Ramanan.
\newblock Detecting activities of daily living in first-person camera views.
\newblock In {\em CVPR}, 2012.

\bibitem{ranzato2014video}
M.~Ranzato, A.~Szlam, J.~Bruna, M.~Mathieu, R.~Collobert, and S.~Chopra.
\newblock Video (language) modeling: a baseline for generative models of
  natural videos.
\newblock {\em arXiv preprint arXiv:1412.6604}, 2014.

\bibitem{rhinehart2018forecast}
N.~Rhinehart and K.~Kitani.
\newblock First-person activity forecasting from video with online inverse
  reinforcement learning.
\newblock {\em TPAMI}, 2018.

\bibitem{rhinehart2016learning}
N.~Rhinehart and K.~M. Kitani.
\newblock Learning action maps of large environments via first-person vision.
\newblock In {\em CVPR}, 2016.

\bibitem{roy2016multi}
A.~Roy and S.~Todorovic.
\newblock A multi-scale cnn for affordance segmentation in rgb images.
\newblock In {\em ECCV}, 2016.

\bibitem{savva2014scenegrok}
M.~Savva, A.~X. Chang, P.~Hanrahan, M.~Fisher, and M.~Nie{\ss}ner.
\newblock Scenegrok: Inferring action maps in 3d environments.
\newblock {\em TOG}, 2014.

\bibitem{sawatzky2017adaptive}
J.~Sawatzky and J.~Gall.
\newblock Adaptive binarization for weakly supervised affordance segmentation.
\newblock {\em ICCV: Workship on Assistive Computer Vision and Robotics}, 2017.

\bibitem{sawatzky2017weakly}
J.~Sawatzky, A.~Srikantha, and J.~Gall.
\newblock Weakly supervised affordance detection.
\newblock In {\em CVPR}, 2017.

\bibitem{selvaraju2017grad}
R.~R. Selvaraju, M.~Cogswell, A.~Das, R.~Vedantam, D.~Parikh, D.~Batra, et~al.
\newblock Grad-cam: Visual explanations from deep networks via gradient-based
  localization.
\newblock In {\em ICCV}, 2017.

\bibitem{springenberg2015striving}
J.~T. Springenberg, A.~Dosovitskiy, T.~Brox, and M.~Riedmiller.
\newblock Striving for simplicity: The all convolutional net.
\newblock {\em ICLR: Workshop}, 2015.

\bibitem{srivastava2015unsupervised}
N.~Srivastava, E.~Mansimov, and R.~Salakhudinov.
\newblock Unsupervised learning of video representations using lstms.
\newblock In {\em ICML}, 2015.

\bibitem{stark2008functional}
M.~Stark, P.~Lies, M.~Zillich, J.~Wyatt, and B.~Schiele.
\newblock Functional object class detection based on learned affordance cues.
\newblock In {\em ICVS}, 2008.

\bibitem{thermos2017deep}
S.~Thermos, G.~T. Papadopoulos, P.~Daras, and G.~Potamianos.
\newblock Deep affordance-grounded sensorimotor object recognition.
\newblock {\em CVPR}, 2017.

\bibitem{villegas2017learning}
R.~Villegas, J.~Yang, Y.~Zou, S.~Sohn, X.~Lin, and H.~Lee.
\newblock Learning to generate long-term future via hierarchical prediction.
\newblock In {\em ICML}, 2017.

\bibitem{vondrick2015anticipating}
C.~Vondrick, H.~Pirsiavash, and A.~Torralba.
\newblock Anticipating the future by watching unlabeled video.
\newblock {\em CVPR}, 2016.

\bibitem{vondrick2016generating}
C.~Vondrick, H.~Pirsiavash, and A.~Torralba.
\newblock Generating videos with scene dynamics.
\newblock In {\em NeurIPS}, 2016.

\bibitem{vondrick2017generating}
C.~Vondrick and A.~Torralba.
\newblock Generating the future with adversarial transformers.
\newblock In {\em CVPR}, 2017.

\bibitem{walker2017pose}
J.~Walker, K.~Marino, A.~Gupta, and M.~Hebert.
\newblock The pose knows: Video forecasting by generating pose futures.
\newblock In {\em ICCV}, 2017.

\bibitem{wang2017binge}
X.~Wang, R.~Girdhar, and A.~Gupta.
\newblock Binge watching: Scaling affordance learning from sitcoms.
\newblock In {\em CVPR}, 2017.

\bibitem{xue2016visual}
T.~Xue, J.~Wu, K.~Bouman, and B.~Freeman.
\newblock Visual dynamics: Probabilistic future frame synthesis via cross
  convolutional networks.
\newblock In {\em NeurIPS}, 2016.

\bibitem{yao2010grouplet}
B.~Yao and L.~Fei-Fei.
\newblock Grouplet: A structured image representation for recognizing human and
  object interactions.
\newblock In {\em CVPR}, 2010.

\bibitem{yao2013discovering}
B.~Yao, J.~Ma, and L.~Fei-Fei.
\newblock Discovering object functionality.
\newblock In {\em ICCV}, 2013.

\bibitem{zhou2016learning}
B.~Zhou, A.~Khosla, A.~Lapedriza, A.~Oliva, and A.~Torralba.
\newblock Learning deep features for discriminative localization.
\newblock In {\em CVPR}, 2016.

\bibitem{zhou2016cascaded}
Y.~Zhou, B.~Ni, R.~Hong, X.~Yang, and Q.~Tian.
\newblock Cascaded interactional targeting network for egocentric video
  analysis.
\newblock In {\em CVPR}, 2016.

\bibitem{zhu2016inferring}
Y.~Zhu, C.~Jiang, Y.~Zhao, D.~Terzopoulos, and S.-C. Zhu.
\newblock Inferring forces and learning human utilities from videos.
\newblock In {\em CVPR}, 2016.

\bibitem{zhu2015understanding}
Y.~Zhu, Y.~Zhao, and S.~Chun~Zhu.
\newblock Understanding tools: Task-oriented object modeling, learning and
  recognition.
\newblock In {\em CVPR}, 2015.

\end{thebibliography}
}

\newpage

\setcounter{section}{0}
\setcounter{figure}{0}
\setcounter{table}{0}
\renewcommand{\thesection}{S\arabic{section}}
\renewcommand{\thetable}{S\arabic{table}}
\renewcommand{\thefigure}{S\arabic{figure}}

\section*{Supplementary Material}

This section contains supplementary material to support the main paper text. The contents include:
\begin{itemize}[leftmargin=*]
\itemsep0em 
    \item (\S\ref{sec:videos}) A video demonstrating our method on clips of human-object interaction.
    \item (\S\ref{sec:ablation}) Ablation study of our model components in \refsec{sec:distill} and \refsec{sec:hotspot_modifications} of the main paper.
    \item (\S\ref{sec:annotation}) EPIC Kitchens data annotation details. 
    \item (\S\ref{sec:l_ant}) Details about the anticipation loss $\mathcal{L}_{ant}$ for EPIC Kitchens described in \refsec{sec:exp} (Datasets).      
    \item (\S\ref{sec:implementation}) Implementation details for our model and experiments in \refsec{sec:grounded-affordances}.
    \item (\S\ref{sec:img2heatmap}) Architecture details for the \SC{Img2heatmap} model introduced in \refsec{sec:grounded-affordances} (Baselines)
    \item (\S\ref{sec:evaluation}) Additional details about the evaluation protocol for our experiments in \refsec{sec:grounded-affordances}.
    \item (\S\ref{sec:examples}) More examples of hotspot predictions on OPRA and EPIC to supplement \reffig{fig:opra-qual} in the main paper.
    \item (\S\ref{sec:clustering}) Clustering visualizations to accompany the results in \refsec{sec:dendo}.
\end{itemize}

\section{Interaction hotspots generated on video clips} \label{sec:videos}
We demonstrate our method on videos from EPIC by computing hotspots for each frame of a video clip. Note that the OPRA test set consists only of static images (\reftbl{tbl:hotspot-eval} in main paper), as following the evaluation protocol in \cite{fang2018demo2vec}. The video can be found on the \href{http://vision.cs.utexas.edu/projects/interaction-hotspots/}{project page}. Our model is trained on all actions, but we show hotspots for 5 frequent actions (cut, mix, adjust, open and wash) for clarity. Our model is able to derive hotspot maps on inactive objects \emph{before} the interaction takes place. During this time, the objects are at rest, and are not being interacted with, yet our model can anticipate how they could be interacted with. In contrast to prior weakly supervised saliency methods, our model generates distinct maps for each action that aligns with object function. Some failure modes include when our model is presented rare objects/actions or unfamiliar viewpoints, for which our model produces diffused or noisy heatmaps.

\begin{table}[t]
\centering
\resizebox{\columnwidth}{!}{
\begin{tabular}{l|ccc|ccc}
\multicolumn{1}{c}{}                &          \multicolumn{3}{c}{OPRA}                    &           \multicolumn{3}{c}{EPIC}                   \\
\cmidrule(l{2pt}r{2pt}){2-4} \cmidrule(l{2pt}r{2pt}){5-7}
                                    & KLD $\downarrow$ & SIM $\uparrow$ & AUC-J $\uparrow$ & KLD $\downarrow$ & SIM $\uparrow$ & AUC-J $\uparrow$ \\
\cmidrule{1-7}
\SC{Ours (basic)}                   & 1.561            & 0.349          & 0.707            & 1.342           & 0.396          & 0.714             \\
\SC{+ Resolution}                   & 1.492            & 0.352          & 0.766            & 1.343           & 0.395          & 0.731             \\ 
\SC{+ L2 pool}                      & 1.489            & 0.349          & 0.770            & 1.361           & 0.385          & 0.727             \\
\SC{+ Anticipation}                 & \B{1.427}        & \B{0.362}      & \B{0.806}        & \B{1.258}       & \B{0.404}      & \B{0.785}         \\
\cmidrule{1-7}
\end{tabular}
}
\caption{\textbf{Ablation study of model enhancements}. Using a larger backbone feature resolution, L2-pooling, and the anticipation module improves our model performance, and are essential for well localized hotspot maps.
} 
\label{tbl:ablation}
\end{table}

\section{Ablation study} \label{sec:ablation}
As noted in the main paper (\refsec{sec:grounded-affordances}), we study the effect of each proposed component in \refsec{sec:distill} and \refsec{sec:hotspot_modifications} on the performance of our model. \reftbl{tbl:ablation} shows how they contribute towards more affordance aware activation maps. Specifically, we see that increasing the backbone resolution to N=28 increases AUC-J from 0.707 to 0.766. Using L2-pooling to ensure that spatial locations produce gradients as a function of their activation magnitudes increases this to 0.770, and using our anticipation model (\refsec{sec:distill} of the main paper) further increases AUC-J to \textbf{0.806}.
Note that within our model, hotspots \emph{could} be be derived at the output hypothesized image, corresponding to hotspots on a regular video frame, but this does not align with the static images, causing a drop in performance (0.806 vs. 0.723 AUC-J). Propagating gradients back through the anticipation module produces correctly aligned gradients. 
More generally, we observe consistent improvement for our components across all metrics on both OPRA and EPIC.

\section{Data collection setup for EPIC-Kitchens annotations} \label{sec:annotation}
As mentioned in \refsec{sec:exp} (Datasets), we collect annotations for interaction keypoints on EPIC Kitchens in order to quantitatively evaluate our method in parallel to the OPRA dataset (where annotations are available). We note that these annotations are collected purely for evaluation, and are not used for training our model. We select the 20 most frequent verbs, and select 31 nouns that afford these interactions. The list of verbs and nouns can be found in \reftbl{tbl:verb_noun_subset}. To select instances for annotation, we first identify frames in the EPIC Kitchen videos which contain the object, and crop out the object using the provided bounding box annotations. Most of these object crops are \emph{active} \ie the objects are not at rest and/or they are being actively manipulated. We discard instances where hands are present in the crop using an off the shelf hand detection model, and manually select for 
\emph{inactive} images from the remaining object crops. These images are then annotated.

\begin{figure}[t!]
\centering
\includegraphics[width=\linewidth]{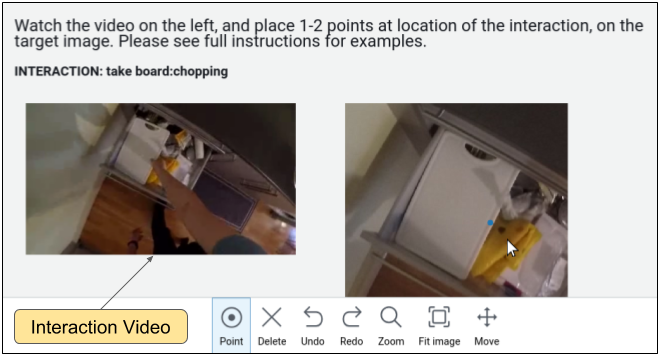}
\caption{\textbf{Interface for collecting annotations on EPIC.} Users are asked to watch a video containing an interaction, and mark keypoints at the location of the interaction.}
\label{fig:mturk}
\end{figure}

We crowdsource these annotations using the Amazon Mechanical Turk platform. Following \cite{fang2018demo2vec}, our annotation setup is as follows. A user is asked to watch a short video (2-5s) of an object interaction (eg. person cutting carrot), and place 1-2 points on an image of the object for where the interaction is performed. We solicited 5 annotations for the same image (from unique users) to account for inter-annotator variance. Overall, we collected 19800 responses from 613 workers for our task, resulting in 1871 annotated (image, verb) pairs. Our annotation interface is shown in \reffig{fig:mturk}.

\begin{figure}[t!]
\centering
\includegraphics[width=\linewidth]{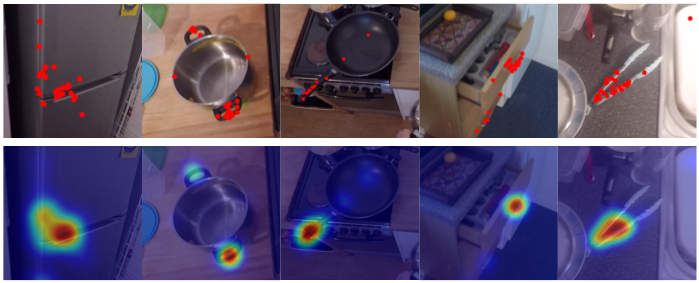}
\caption{\textbf{Top:} Some example annotations provided as keypoints on EPIC object crops \textbf{Bottom:} The resulting heatmaps derived from these annotations, to be used as ground truth for evaluation.}
\label{fig:annotations}
\end{figure}

Finally, following~\cite{fang2018demo2vec}, we convert these annotations into a heatmap by centering a gaussian distribution over each point. We use this heatmap as our ground truth $\mathcal{M}$ described in \refsec{sec:exp} (Datasets). Some examples of these collected annotations and their correspondingly derived heatmaps are shown in \reffig{fig:annotations}.

\begin{table}[t!]
\centering
\resizebox{\columnwidth}{!}{
\begin{tabular}{|p{1cm}|p{8cm}|} \hline
\B{verbs} & take, put, open, close, wash, cut, mix, pour, throw, move, remove, dry, turn-on, turn, shake, turn-off, peel, adjust, empty, scoop \\ \hline
\B{nouns} & board:chopping, bottle, bowl, box, carrot, colander, container, cup, cupboard, dishwasher, drawer, fork, fridge, hob, jar, kettle, knife, ladle, lid, liquid:washing, microwave, pan, peeler:potato, plate, sausage, scissors, spatula, spoon, tap, tongs, tray \\ \hline
\end{tabular} 
}
\caption{Verbs and nouns annotated for EPIC.}
 \label{tbl:verb_noun_subset}
\end{table}

Compared to asking annotators to label static images with affordance labels (\eg, label "openable"), annotations collected by watching videos and then placing points is well-aligned with our objective of learning fine-grained object interaction. The annotations are better localized and are grounded in real human demonstration, making them meaningful for evaluation.

\section{Anticipation loss for EPIC-Kitchens} \label{sec:l_ant}
As mentioned in \refsec{sec:distill} and \refsec{sec:hotspot_modifications}, we require inactive images to train the anticipation model. For OPRA, these are the catalog photos of the object provided with each video. In EPIC, we crop out inactive objects from frames using the provided bounding boxes $\mathcal{B}$, and select the inactive image that has the same class label as the object in the video. Unlike OPRA, these images may be visually different from the object in the video, preventing us from using the L2 loss directly. Instead, we use a triplet loss for the anticipation loss term as follows:
\begin{equation*}
    \begin{split}
    \mathcal{L}^\prime_{ant}(\mathcal{V}, x_{I}, x_{I'}) =  \max \left [
        0, d(P(x_{t^*}), P(\widetilde{x}_I) \right. \\
        \left. - d(P(x_{t^*}), P(\widetilde{x}_{I'})) + M \right ],
    \end{split}
\label{eq:triplet}
\end{equation*}
where $x_I$ and $x_{I'}$ represent inactive image features with the correct and incorrect object class respectively. $d$ denotes Euclidean distance, and $M$ is the margin value. We normalize the inputs before computing the triplet loss, thus we keep the margin value fixed at 0.5. This term ensures that inactive objects of the correct class can anticipate active features better than incorrect classes, and is less sensitive to appearance mismatches compared to \refeqn{eqn:anticipation} in the main paper.

\begin{figure*}[t!]
\centering
\includegraphics[width=\linewidth]{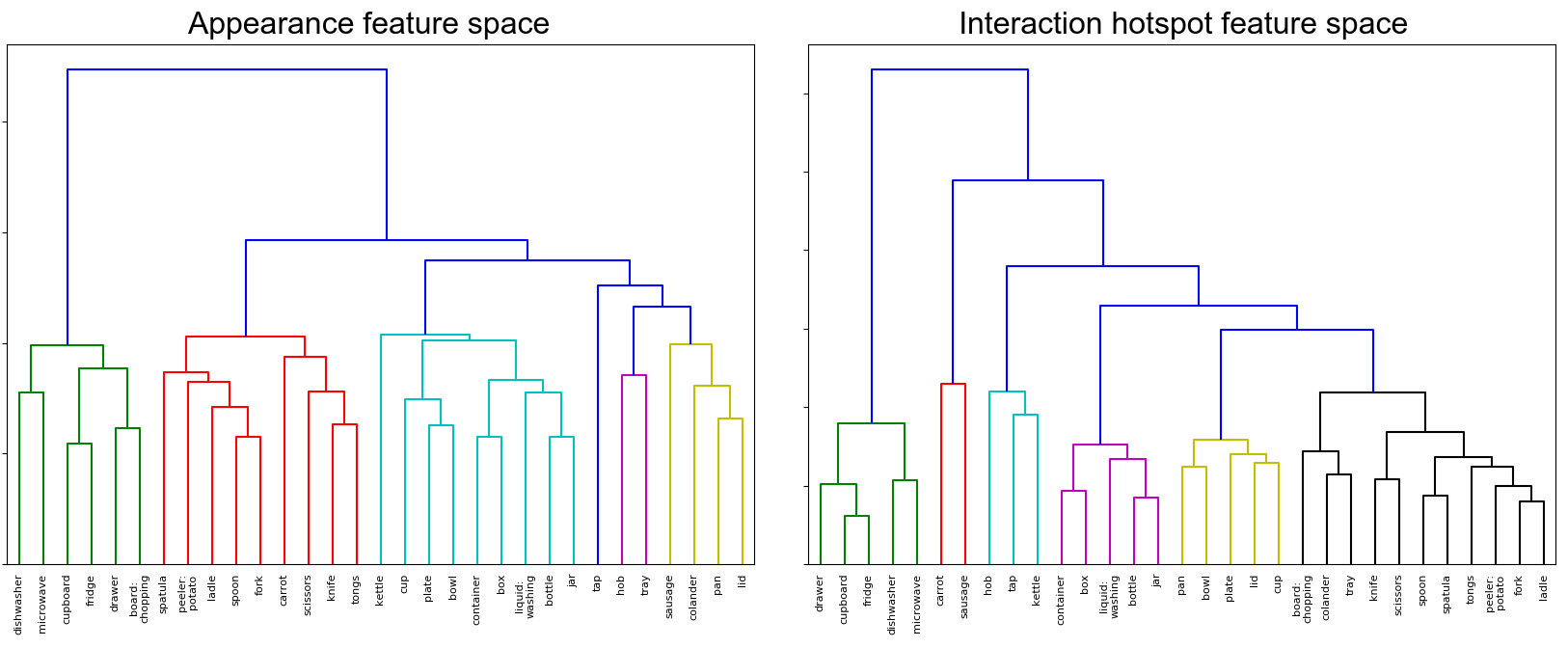}
\caption{\textbf{Clustering of the average object embedding in the appearance vs. interaction hotspot feature space.} Appearance features capture similarities in shapes and textures (knife, tongs) and object co-occurrence (pan, lid; cup, kettle). In contrast, our interaction hotspot features encode similarities that are aligned with object function and use (cupboard, microwave, fridge - characteristically swung open; carrot, sausage - cut and held similarly). Similarity in this space refers to L2 distance between average object embeddings.}
\label{fig:dendograms}
\end{figure*}

\section{Implementation details} \label{sec:implementation}
We provide implementation and training details for our experiments in  \refsec{sec:exp}. For all experiments, we use an ImageNet~\cite{krizhevsky2012imagenet} pretrained ResNet-50~\cite{he2016deep} modified for higher output resolution. To increase the output dimension from $n = 7 $ to $n = 28$, we set the spatial stride of res$_4$ and res$_5$ to 1 (instead of 2), and use dilation of 2 (res$_4$) and 4 (res$_5$) for its filters to preserve the original scale.  For $\mathcal{F}_{ant}$ we use 2 sets of  (conv-bn-relu) blocks, with 3x3 convolutions, maintaining the channel dimension $d=2048$. We use a single layer LSTM (hidden size 2048) as $\mathbb{A}$, and train using chunks of 16 frames at a time. For our combined loss (\refeqn{eqn:total_loss} in the main paper), we set $\lambda_{cls} = \lambda_{aux} = 1$ and $\lambda_{ant} = 0.1$ based on validation experiments. Our models are implemented in PyTorch. Adam with learning rate 1e-4, weight decay 5e-4 and batch size $128$ is used to optimize the models parameters.

\section{Architecture details for supervised baselines} \label{sec:img2heatmap}
The \SC{Img2heatmap} model in \refsec{sec:grounded-affordances} is a fully convolutional encoder-decoder to predict the affordance heatmap for an image. The encoder is an ImageNet pretrained VGG16 backbone (up to conv5), resulting in an encoded feature with 512 channels and spatial extent 7. This feature is passed through a decoder with an architecture mirroring the backbone, where the max-pooling operations are replaced with bilinear upsampling operations. This results in an output of the same dimension as the input, and as many channels as the number of actions. The output of this network is fed through a sigmoid operator and reconstruction loss against the ground truth affordance heatmap is calculated using binary cross-entropy.

\section{Evaluation protocol for grounded affordance prediction} \label{sec:evaluation}
As discussed in \refsec{sec:grounded-affordances}, the heatmaps generated by our model and the baselines are evaluated against the manually annotated ground truth heatmaps provided in the OPRA dataset and collected on EPIC (results in \reftbl{tbl:hotspot-eval}). For a single action (\eg ``press" a button), the ground truth heatmaps may occur distributed across several instances (\eg different clips of people pressing different buttons on the same object). We simply take the union of all these heatmaps as our target affordance heatmap for the action. For evaluation, this leaves us with 1042 (image, action) pairs in OPRA, and 571 (image, action) pairs in EPIC. For AUC-J, we binarize heatmaps using a threshold of $0.5$ for evaluation.

\section{Additional examples of generated hotspot maps} \label{sec:examples}
\begin{figure*}[t!]
\centering
\includegraphics[width=\linewidth]{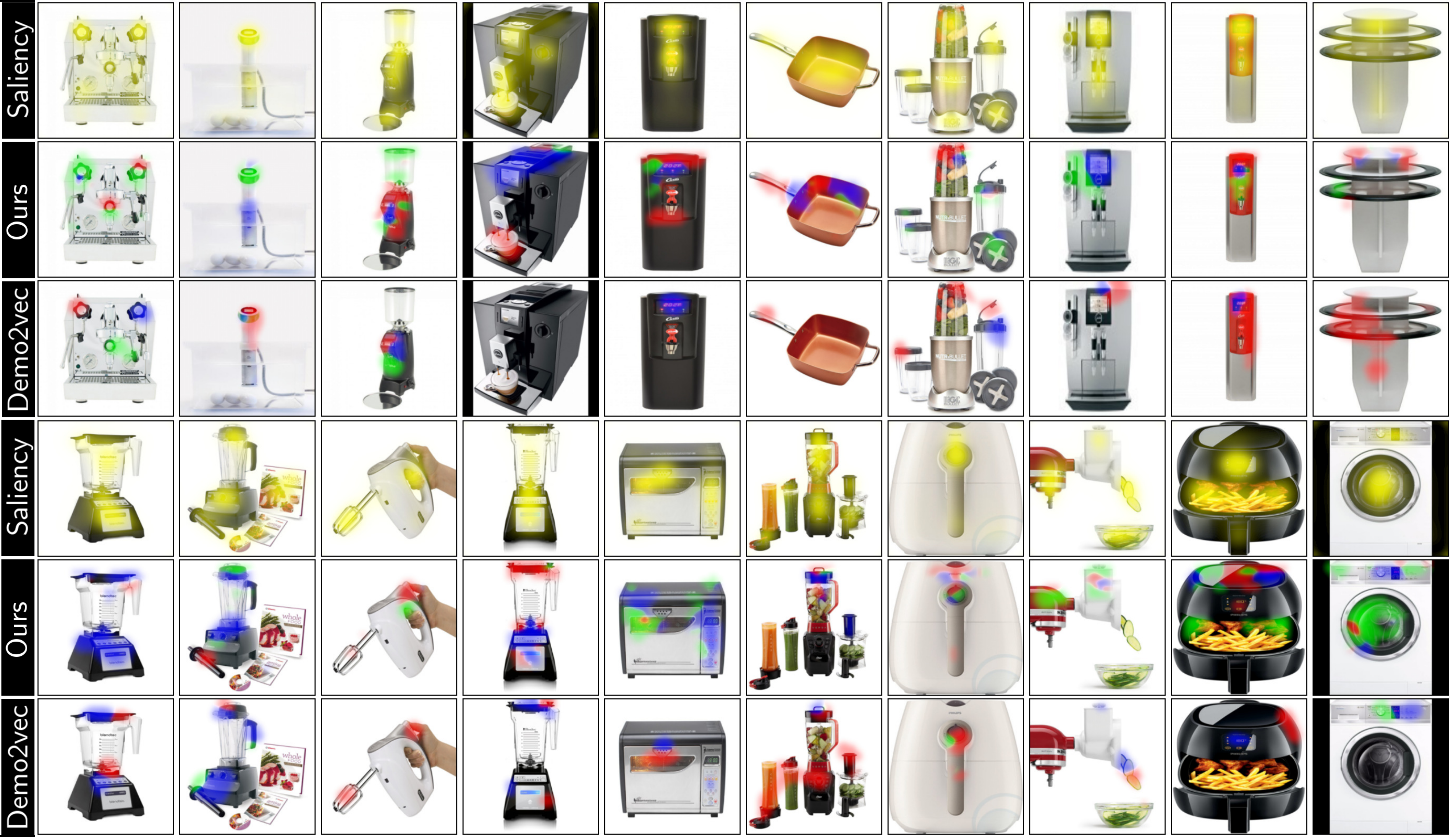}
\caption{\textbf{Generated affordance heatmaps on inactive images from OPRA.} Our interaction hotspot maps show holdable, rotatable, and pushable regions (in red, green, and blue respectively). Saliency heatmaps do not discriminate between interactions and produce a single heatmap shown in yellow. Recall that the \SC{demo2vec} approach~\cite{fang2018demo2vec} is strongly supervised, whereas our approach is weakly supervised. Some failure cases due to small or unfamiliar objects can be seen in the last 4 examples.}
\label{fig:opra-supp}
\end{figure*}

\begin{figure*}[t!]
\centering
\includegraphics[width=\linewidth]{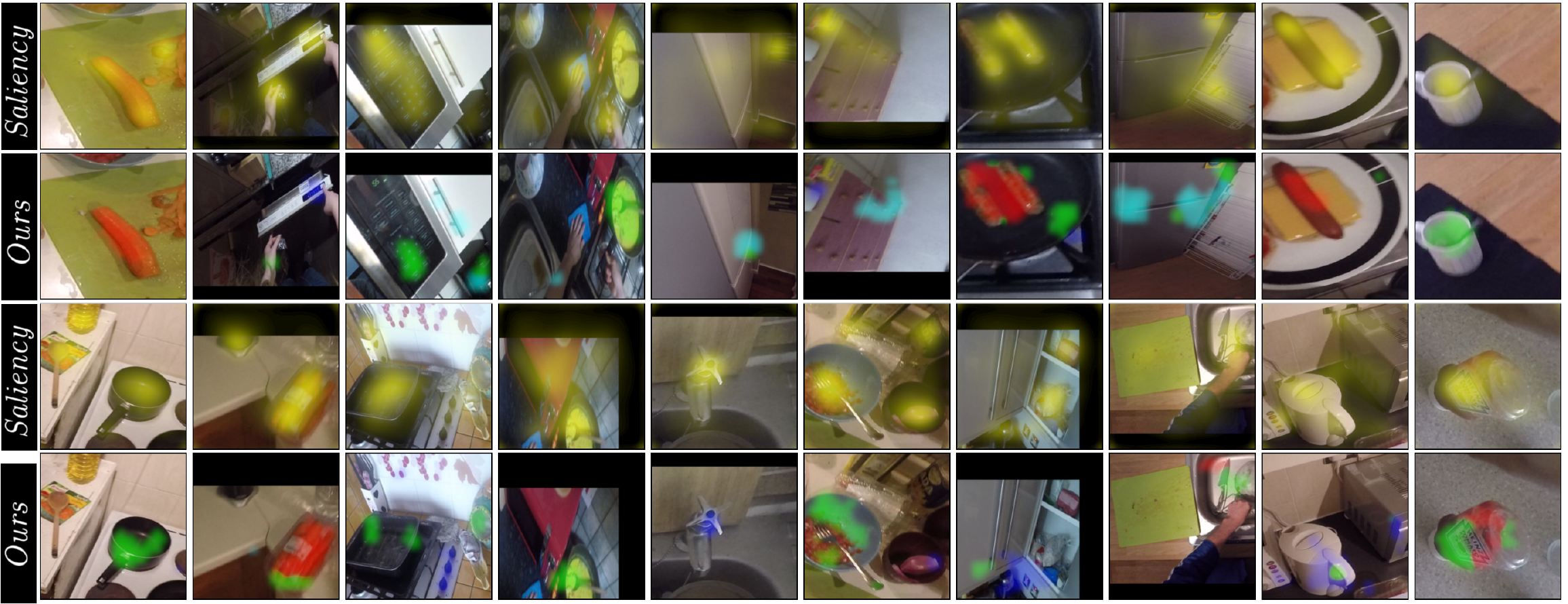}
\caption{\textbf{Generated interaction hotspot maps on inactive images from EPIC-Kitchens.} Our interaction hotspot maps show cuttable, mixable, adjustable, and openable regions (in red, green, blue, and cyan, respectively). Failure cases can be seen in the last 4 examples.}
\label{fig:epic-supp}
\end{figure*}

We provide more examples of our hotspot maps to accompany our results in the main paper. \reffig{fig:opra-supp} and \reffig{fig:epic-supp} contains more examples of these on OPRA and EPIC respectively to supplement our results in \reffig{fig:opra-qual} in the main paper. Unlike the baselines, our model highlights multiple distinct affordances for an object and does so without heatmap annotations during training. The last 4 images in each figure show some failure cases where our model is unable to produce heatmaps for small or unfamiliar objects. 

\section{Clustering visualization for appearance vs. our interaction hotspot features} \label{sec:clustering}
We show the full clustering of objects in the appearance vs. interaction hotspot feature space to supplement the nearest neighbor visualizations presented in \refsec{sec:dendo} of the main paper. Each object is represented by a vector obtained by averaging the embeddings of all instances for that specific object class. The resultant \emph{average object representations} for all classes are then clustered using agglomerative hierarchical clustering. L2 distance in this space represents average similarity between object classes. 

\reffig{fig:dendograms} shows how our learned representation groups together objects related by their function and interaction modality, more so than the original appearance-based visual representation. Appearance features capture similarities in shapes and textures (knife, tongs) and object co-occurrence (pan, lid; cup, kettle). In contrast our representation encodes object function. Cupboards, microwaves, fridges that are characteristically swung opened; knives and scissors that afford the same cutting action; carrots, sausages that are cut and held in the same manner, are clustered together.

\end{document}